\documentclass[]{fairmeta}
\usepackage{xspace}
\usepackage{amsmath}
\usepackage{amssymb}
\usepackage{tabularx}
\usepackage{adjustbox}
\usepackage{colortbl}
\usepackage{booktabs}
\usepackage{multirow}
\usepackage{enumitem}


\definecolor{titlecolor}{HTML}{8F1329}
\title{\textcolor{titlecolor}{RadioWorld}:\hspace{0.18cm} Transferring World Models\\ to Radio-Frequency Radiance Fields}

\abstract{
Electromagnetic (EM) wave propagation in the physical world constitutes a Radio-Frequency (RF) radiance field, providing rich information for 6G communication and sensing. However, existing neural RF field methods suffer from three fundamental limitations: (1) they predominantly rely on per-scene optimization and cannot generalize to unseen scenes; (2) they require costly reference RF measurements in each target scene; and (3) they degrade severely when extrapolating beyond the vicinity of reference measurements. To overcome these limitations, we present \textbf{RadioWorld}, a general framework that adapts a world model to synthesize RF radiance fields directly from physical scene geometry and transceiver locations, without requiring any reference RF measurements at inference. Specifically, RadioWorld couples two specialized modules within a dual-stream Mixture-of-Transformers (MoT) architecture: a \emph{Scene-Understanding Transformer} that encodes 3D scene geometry and transceiver locations to extract environment-aware global EM propagation features, and a \emph{Field-Generation Transformer} that attends to these features to synthesize the RF radiance field with the local query conditions. We also establish \textbf{UrbanRF-Bench}, a large-scale ray-traced benchmark spanning $657$ urban scenes and $32{,}850$ RF radiance fields across seven global cities. Extensive experiments demonstrate that RadioWorld significantly outperforms specialized baselines and accurately synthesizes RF fields at arbitrary locations in unseen cities. Code, datasets, and checkpoints will be released upon acceptance.
}

\author[1]{Hu Zhou}
\author[2]{Jiazhou Zhou}
\author[1]{Shuowen Zhang}
\author[3]{Ying-Chang Liang}
\author[1]{Liang Liu}

\affiliation[1]{The Hong Kong Polytechnic University}
\affiliation[2]{The Hong Kong University of Science and Technology (Guangzhou)}
\affiliation[3]{University of Electronic Science and Technology of China}

\metadata[Project page]{\url{https://radioworld-ai.github.io/}}
\correspondence{\email{hu-eee.zhou@polyu.edu.hk, liang-eie.liu@polyu.edu.hk}}

\begin{document}

\maketitle

\begin{center}
    \includegraphics[width=\textwidth]{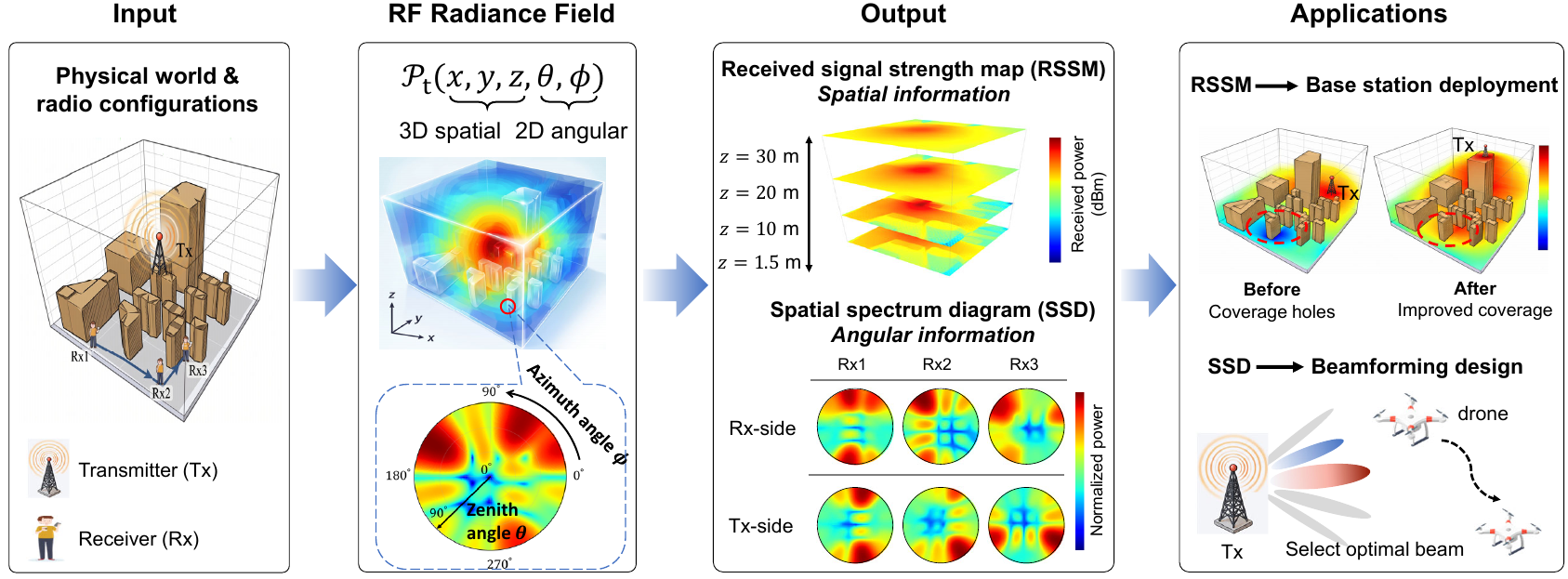}
    \captionof{figure}{\textbf{RadioWorld} is a feed-forward world model that synthesizes received signal strength maps (RSSMs) at different heights and spatial spectrum diagrams (SSDs) at arbitrary locations from 3D scene geometry and transceiver locations, without any reference RF measurements, empowering 6G communication and sensing.}
    \label{fig:overview}
\end{center}

\section{Introduction}
\vspace{3mm}
\label{sec:intro}

Simulating electromagnetic (EM) wave interactions with 3D environments is fundamental for understanding the physical world, which has wide applications in wireless digital twins~\cite{orekondy2023winert,hehn2025wigatr}, such as base station deployment and directional beamforming in the low-altitude economy (Fig.~\ref{fig:overview}). However, this remains highly challenging because EM waves undergo intricate reflection, diffraction, scattering, and refraction across 3D urban environments~\cite{goldsmith2005wireless,yun2015ray}. From a radiometric perspective, EM wave propagation forms a \textbf{5D spatio-angular radio-frequency (RF) radiance field} representing the spatial distribution of angular RF power across 3D spaces and 2D radiation directions~\cite{mildenhall2020nerf,zhao2023nerf2,yang2026graf}. Projecting this 5D field onto spatial planes and angular domains yields received signal strength maps (RSSMs) for coverage evaluation and spatial spectrum diagrams (SSDs) for directional beamforming in 6G networks~\cite{giordani2019beam}, respectively.

Conventionally, ray-tracing solvers are used for simulating the RF radiance field~\cite{yun2015ray,hoydis2022sionna,ait_aoudia2025sionnart}, but are of prohibitive cost when dealing with large-scale urban environments. To accelerate the simulation, learning-based surrogates have evolved along three directions. First, some works~\cite{levie2021radiounet,wang2025radiodiff,jia2025radiomamba} predict RSSMs from scene geometry, but discard directional angular information. Second, to capture the directional information, neural RF radiance field-based methods adapt NeRF~\cite{zhao2023nerf2,lu2023newrf} or 3D Gaussian Splatting (3DGS)~\cite{wen2025wrfgs,wen2025wrfgsplus,yang2025gsrf} for SSD synthesis, but require per-scene optimization. Third, only a few works~\cite{yang2026graf,wen2026point2radio} attempt to synthesize SSDs across different scenes, but still rely on target-scene reference RF measurements or are restricted to simple indoor layouts. Overall, existing methods still suffer from three key challenges: (1) reliance on target-scene reference RF measurements, (2) severe performance degradation when extrapolating beyond reference measurement regions, and (3) inability to generalize across complex outdoor cities without per-scene retraining.

Recent advances in world models~\cite{nvidia2026cosmos3,mur2026v,hafner2023dreamerv3,bruce2024genie,yang2024unisim,brooks2024sora,nvidia2025cosmos,bar2025nwm,assran2025vjepa2} have offered a promising solution to these challenges. Pretrained on large-scale video generation and next physical state prediction tasks, world models internalize strong perceptual and generative priors over 3D spatial topology and optical light transport. Fortunately, both optical light transport and RF signal propagation are governed by Maxwell's equations and share fundamental propagation laws, such as blockages and specular reflections. However, different wavelengths of optical light and RF signals introduce distinct propagation properties, e.g., the longer wavelength of RF signals introduces rich non-line-of-sight (NLoS) multipath propagation that are less prominent for optical light. These similarities and differences motivate a fundamental question: \emph{Can the spatial understanding and physical modeling capabilities of world models be transferred beyond optical light to synthesize invisible RF radiance fields from the physical world?}

To this end, we propose \textbf{RadioWorld}, a unified framework that adapts a pretrained Mixture-of-Transformers (MoT) world model~\cite{liang2024mot,nvidia2026cosmos3} to synthesize RF radiance fields from 3D scene geometry and transceiver configurations alone (Fig.~\ref{fig:overview}). Specifically, a \emph{Scene-Understanding Transformer} encodes scene geometry and transceiver locations into environment-aware global EM propagation features (while supporting optional text-based link-level propagation reasoning), and a \emph{Field-Generation Transformer} attends to these features alongside local RSSM/SSD query conditions to synthesize multi-height RSSMs and bidirectional SSDs (as well as optional multi-frame SSDs). Furthermore, since RSSM and SSD under a given scene and transceiver configuration are deterministic whereas world models rely on iterative flow matching~\cite{lipman2023flow}, we replace multi-step denoising with \emph{single-step deterministic regression}, eliminating sampling variance and achieving up to $2.5\times$ faster inference with higher fidelity.

To train and evaluate RadioWorld, we establish \textbf{UrbanRF-Bench}, a ray-traced benchmark of $657$ urban scenes and $32{,}850$ RF radiance fields across seven global cities, providing multi-height RSSMs, bidirectional SSDs (Tx- and Rx-side), multi-frame SSDs, and link-level propagation descriptions. Our contributions are three-fold:
\begin{itemize}
    \item We propose \textbf{RadioWorld}, a feed-forward world-model framework that synthesizes multi-height RSSMs and bidirectional SSDs directly from the physical world and transceiver locations, requiring zero reference RF measurements at inference.
    \item We establish \textbf{UrbanRF-Bench}, a large-scale ray-traced benchmark spanning $657$ scenes across seven cities under a standardized cross-city evaluation protocol.
    \item Extensive experiments show that \textbf{RadioWorld} outperforms specialized baselines across all benchmarks, generalizes zero-shot to arbitrary locations in unseen cities and enhances the performance of downstream tasks.
\end{itemize}

\section{Related Work}
\label{sec:related_work}

\paragraph{RSSM Synthesis of RF Radiance Fields.}
RSSM synthesis methods translate the scene layout and transmitter location into a received power map~\cite{romero2022radiomap,yapar2023challenge,lee2023pmnet,zhang2023rmegan,levie2021radiounet,wang2025radiodiff,jia2025radiomamba,zhu2026df,zhu2026video}. For example, RadioUNet~\cite{levie2021radiounet} uses U-Nets, demonstrating the feasibility of using neural networks for fast coverage estimation. RadioDiff~\cite{wang2025radiodiff} formulates RSSM construction as conditional diffusion generation. RadioMamba~\cite{jia2025radiomamba} introduces a hybrid state-space U-Net architecture to balance prediction accuracy and computational efficiency. Although these methods serve as strong RSSM synthesis baselines, their output is restricted to scalar power, while failing to model the directional power distributions over angular domains.

\paragraph{SSD Synthesis of RF Radiance Fields.}
NeRF$^2$~\cite{zhao2023nerf2} and NeWRF~\cite{lu2023newrf} predict an RF radiance field~\cite{mildenhall2020nerf} based on the reference SSD measurements per scene. While WRF-GS~\cite{wen2025wrfgs}, WRF-GS+~\cite{wen2025wrfgsplus}, and GSRF~\cite{yang2025gsrf} replace NeRF with 3D Gaussian Splatting (3DGS)~\cite{kerbl2023gaussian} for faster inference, they all need per-scene optimization. Furthermore, GRaF~\cite{yang2026graf} generalizes across scenes but is still conditioned on reference SSD measurements. Point2Radio~\cite{wen2026point2radio} predicts from scene layouts but trains small networks from scratch on limited indoor layouts. Different from them, RadioWorld focuses on the more diverse and complex outdoor urban environments and synthesizes the SSD in a feed-forward pass given any scene geometry and arbitrary transceiver locations.

\paragraph{World Models.}
Large-scale visual pretraining yields generalizable representations that can be effectively transferred to dense spatial prediction~\cite{oquab2023dinov2,tschannen2025siglip2,simeoni2025dinov3}. Besides, generative diffusion priors have similarly been repurposed for geometric targets such as depth estimation via either multi-step~\cite{ke2024marigold,fu2024geowizard} or single-step~\cite{garcia2025finetuning,he2025lotus} inference. Building upon these representation and generation advances, recent world models learn the underlying physical laws and spatiotemporal dynamics of complex scenes from massive video and embodied AI data~\cite{hafner2023dreamerv3,bruce2024genie,yang2024unisim,brooks2024sora,nvidia2025cosmos,bar2025nwm,assran2025vjepa2,mur2026v}. Such large-scale pretraining empowers world models with rich visual priors, strong 3D spatial perception and reasoning capabilities, laying the foundation for understanding the physical world and EM wave propagation.

\section{Method}
\label{sec:method}

\subsection{Preliminaries and Problem Definition}
\label{sec:problem}

We discretize each 3D urban scene into $X \times Y \times Z$ grids at $1\,\mathrm{m}$ spatial resolution, represented in bird's-eye view (BEV) by a binary building-occupancy map $\mathbf{O}\in\{0,1\}^{X \times Y}$ and a building-height map $\mathbf{H}\in\mathbb{R}^{X \times Y}$. Let $\mathbf{t}=(x_t,y_t,z_t)$ and $\mathbf{r}=(x_r,y_r,z_r)$ denote the 3D coordinates of the transmitter (Tx) and receiver (Rx), respectively. EM wave propagation induced by a Tx at $\mathbf{t}\in\mathbb{R}^3$ is described as a \textbf{5D spatio-angular RF radiance field} $\mathcal{P}_{\mathbf{t}}(\mathbf{p}, \boldsymbol{\omega}) \in \mathbb{R}_{\ge 0}$, which characterizes the angular power density arriving at any 3D spatial coordinate $\mathbf{p}=(x,y,z)\in\mathbb{R}^3$ from any spherical direction $\boldsymbol{\omega}=(\theta,\phi)\in\mathbb{S}^2$ (with zenith angle $\theta$ and azimuth angle $\phi$). Projecting this 5D field onto spatial planes and angular domains yields two physical quantities for wireless communication and sensing:

\paragraph{Received Signal Strength Map (RSSM).}
For a Tx at $\mathbf{t}$, integrating $\mathcal{P}_{\mathbf{t}}(\mathbf{p}, \boldsymbol{\omega})$ over all directions $\boldsymbol{\omega}\in\mathbb{S}^2$ across a horizontal plane at query height $z_{q}$ yields the spatial RSSM $\mathbf{R}_{z_{q}} \in \mathbb{R}^{X \times Y}$:
\begin{equation}
    \mathbf{R}_{z_q}(x,y) = \int_{\mathbb{S}^2} \mathcal{P}_{\mathbf{t}}\big(\mathbf{p}=(x,y,z_q), \boldsymbol{\omega}\big)\,\mathrm{d}\boldsymbol{\omega},
    \label{eq:pcm_projection}
\end{equation}
which characterizes signal coverage at elevation $z_q$ and can be used for network coverage planning and optimization.

\paragraph{Spatial Spectrum Diagram (SSD).}
To characterize bidirectional SSDs, we define a forward RF radiance field $\mathcal{P}_{\mathbf{t}}(\mathbf{p},\boldsymbol{\omega})$, generated by the Tx at $\mathbf{t}$, and a reverse RF radiance field $\mathcal{P}_{\mathbf{r}}(\mathbf{p},\boldsymbol{\omega})$, generated by a virtual Tx placed at the Rx position $\mathbf{r}$. Both fields are defined over the same propagation environment. Under channel reciprocity, the reverse field captures the reciprocal propagation paths of the forward field, with the propagation directions reversed. Slicing $\mathcal{P}_{\mathbf{t}}$ at the Rx coordinate along the upper hemisphere yields the Angle-of-Arrival (AoA) spectrum $\mathbf{S}^{\mathrm{Rx}}\in\mathbb{R}^{90\times 360}$. Similarly, slicing the reverse field $\mathcal{P}_{\mathbf{r}}$ at the Tx coordinate along the lower hemisphere yields the Angle-of-Departure (AoD) spectrum $\mathbf{S}^{\mathrm{Tx}}\in\mathbb{R}^{90\times 360}$:
\begin{align}
    \mathbf{S}^{\mathrm{Rx}}(m,n) &= \mathcal{P}_{\mathbf{t}}\big(\mathbf{p}=\mathbf{r}, \boldsymbol{\omega}=(\theta_m, \phi_n)\big), \label{eq:pas_rx_projection} \\
    \mathbf{S}^{\mathrm{Tx}}(m,n) &= \mathcal{P}_{\mathbf{r}}\big(\mathbf{p}=\mathbf{t}, \boldsymbol{\omega}=(\theta_m, \phi_n)\big), \label{eq:pas_tx_projection}
\end{align}
where $m=1,\ldots,90$ indexes zenith angles at $1^\circ$ resolution over the upper hemisphere ($\theta_m \in [0^\circ, 90^\circ)$) for $\mathbf{S}^{\mathrm{Rx}}$ and the lower hemisphere ($\theta_m \in [90^\circ, 180^\circ)$) for $\mathbf{S}^{\mathrm{Tx}}$, and $n=1,\ldots,360$ indexes azimuth angles at $1^\circ$ resolution over $[0^\circ, 360^\circ)$, as measured in practice by Bartlett beamforming with a planar array~\cite{vantrees2002optimum}. In practice, SSDs can be used for AoA/AoD estimation and beam selection in wireless sensing.

\paragraph{Problem Definition.}
Without requiring any reference RF measurements at inference time, our goal is to learn a unified feed-forward model $f_\theta$ that directly synthesizes the spatial RSSM and bidirectional SSD from scene geometry $(\mathbf{O}, \mathbf{H})$, transceiver coordinates $\{\mathbf{t}, \mathbf{r}\}$, and query height $z_q$:
\begin{equation}
    \big(\mathbf{R}_{z_q},\, \mathbf{S}^{\mathrm{Rx}},\, \mathbf{S}^{\mathrm{Tx}}\big) = f_\theta\big(\mathbf{O}, \mathbf{H}, \mathbf{t}, \mathbf{r}, z_q\big).
    \label{eq:prediction_tasks}
\end{equation}
In practice, rather than jointly outputting all three tensors in a single forward pass, $f_\theta$ is queried on demand according to the target query mode $q$ (i.e., an RSSM at height $z_q$ or a directional SSD $\mathbf{S}^{\mathrm{Rx}} / \mathbf{S}^{\mathrm{Tx}}$) through the unified token interface summarized in Tab.~\ref{tab:query_slots}.

\subsection{Dual-Stream World Model Architecture}
\label{sec:architecture}

\begin{figure}[t]
  \centering
  \includegraphics[width=\textwidth]{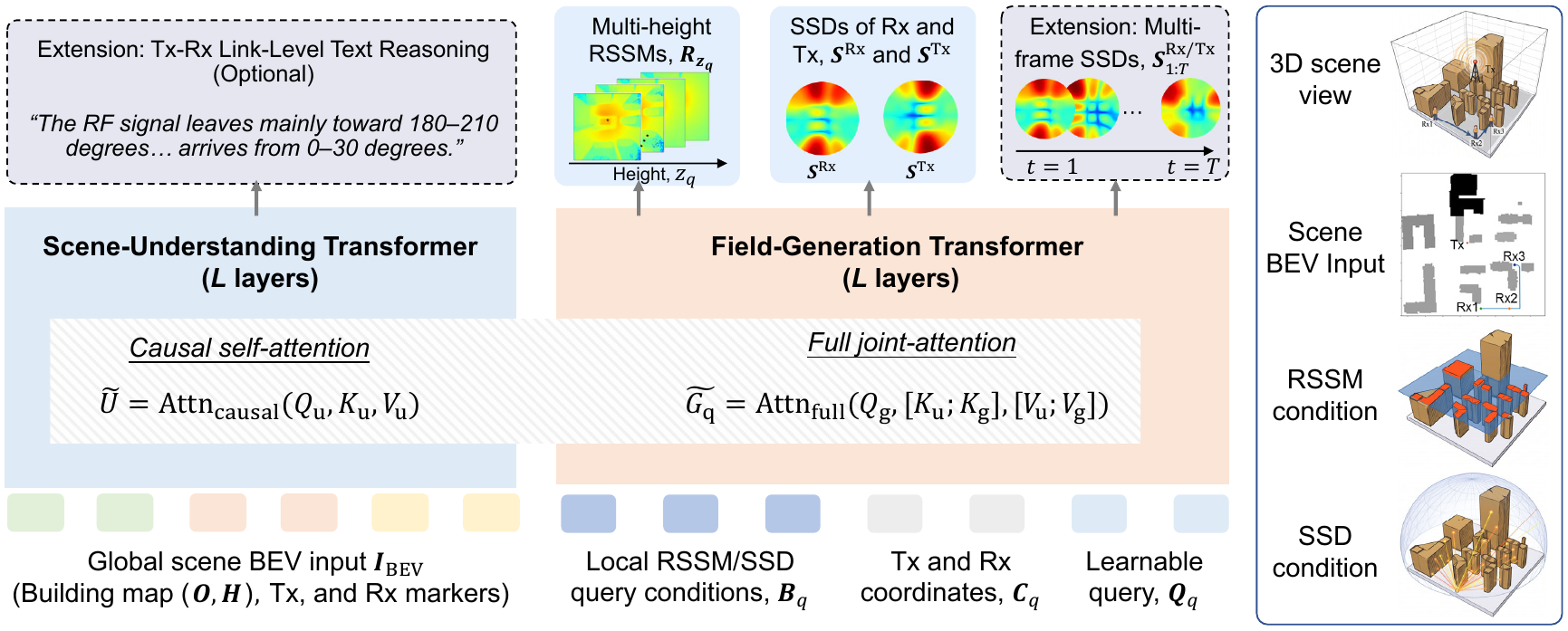}
  \caption{\textbf{Overview of RadioWorld.} The dual-stream world model couples a \emph{Scene-Understanding Transformer} that encodes a global BEV representation $\mathbf{I}_{\mathrm{bev}}$ into environment-aware propagation features $\mathbf{U}$, and a \emph{Field-Generation Transformer} that performs joint attention between the two streams while processing local RSSM/SSD query conditions $\mathbf{B}_q$, coordinate tokens $\mathbf{C}_q$, and learnable queries $\mathbf{Q}_q$ to synthesize RSSMs and SSDs. Inherited from the VLM and video world-model backbone, the architecture naturally supports link-level text propagation reasoning and multi-frame SSD synthesis as extensions.}
  \label{fig:arch}
\end{figure}

\paragraph{Motivation for Dual-Stream Modeling.}
RadioWorld is designed to support two complementary forms of wireless understanding. The first is RF field synthesis, which predicts RSSMs and SSDs, while the second is link-level RF propagation reasoning, such as its dominant arrival and departure directions. Although these tasks share the same scene geometry and RF propagation context, they have fundamentally different output structures: dense RF fields versus variable-length textual sequences. To accommodate these heterogeneous outputs, we thus adopt a dual-stream Mixture-of-Transformers (MoT) architecture comprising two Transformers with separate weights (Fig.~\ref{fig:arch}), where a \emph{Scene-Understanding Transformer} is initialized from a vision-language model (VLM), and a \emph{Field-Generation Transformer} operates in the latent space of the frozen Wan~2.2 VAE~\cite{wang2025wan} ($\mathcal{E}, \mathcal{D}$ denote the encoder and decoder of the VAE).

\paragraph{Scene-Understanding Transformer.}
To capture global EM propagation context from 3D geometry and transceiver locations, we encode the urban layout and Tx/Rx locations into a six-channel BEV raster $\mathbf{I}_{\mathrm{bev}}\in\mathbb{R}^{6\times X\times Y}$. A lightweight adapter maps $\mathbf{I}_{\mathrm{bev}}$ into the frozen SigLIP~2 encoder~\cite{tschannen2025siglip2} to produce initial spatial tokens $\mathbf{U}^{(0)}$. The understanding stream processes $\mathbf{U}^{(0)}$ with causal self-attention across $L$ layers, yielding hidden states $\mathbf{U}=\{\mathbf{U}^{(l)}\}_{l=1}^{L}$ that encode global EM propagation features while providing link-level RF propagation reasoning capabilities (Sec.~\ref{sec:extensions}).

\paragraph{Field-Generation Transformer.}
To synthesize a target field at a specific query manifold and location $q$, the generation stream processes a concatenated token sequence in the VAE latent space:
\begin{equation}
    \mathbf{G}_q = [\mathbf{B}_q;\, \mathbf{C}_q;\, \mathbf{Q}_q],
    \label{eq:gen_seq}
\end{equation}
where $\mathbf{B}_q$ denotes \emph{target-aligned geometric condition tokens} encoded from a local query map $\mathcal{G}_q$ (Sec.~\ref{sec:conditions}) via the frozen VAE encoder $\mathcal{E}$; $\mathbf{C}_q$ is a Fourier-encoded~\cite{tancik2020fourier} coordinate-and-type token specifying transceiver locations and query type; and $\mathbf{Q}_q$ denotes learnable query tokens initialized on the target latent grid.

\paragraph{Cross-Stream Joint Attention.}
The two streams interact exclusively through layer-wise attention. At layer $l$, projecting the incoming states $\mathbf{U}^{(l-1)}$ and $\mathbf{G}_q^{(l-1)}$ ($\mathbf{G}_q^{(0)}=\mathbf{G}_q$) gives per-stream $(\mathbf{Q}_{\mathrm{u}},\mathbf{K}_{\mathrm{u}},\mathbf{V}_{\mathrm{u}})$ and $(\mathbf{Q}_{\mathrm{g}},\mathbf{K}_{\mathrm{g}},\mathbf{V}_{\mathrm{g}})$, updated via the following attention mechanisms:
\begin{align}
    \tilde{\mathbf{U}} &= \mathrm{Attn}_{\mathrm{causal}}\big(\mathbf{Q}_{\mathrm{u}},\mathbf{K}_{\mathrm{u}},\mathbf{V}_{\mathrm{u}}\big), \label{eq:attn_und}\\
    \tilde{\mathbf{G}}_q &= \mathrm{Attn}_{\mathrm{full}}\big(\mathbf{Q}_{\mathrm{g}},[\mathbf{K}_{\mathrm{u}};\mathbf{K}_{\mathrm{g}}],[\mathbf{V}_{\mathrm{u}};\mathbf{V}_{\mathrm{g}}]\big). \label{eq:attn_gen}
\end{align}
Generation tokens attend to global scene context at every layer, whereas the understanding stream remains causal.

\paragraph{Single-Step Deterministic Readout.}
\label{sec:readout}
Pretrained video world models synthesize images/videos by iteratively integrating a learned velocity $v_\theta$ from random noise at $\sigma{=}1$ via flow matching~\cite{lipman2023flow,liu2023rectified,peebles2023dit}. Since RSSM and SSD under fixed scene geometry and transceiver locations are deterministic, stochastic sampling introduces variance and latency. We therefore replace random noise with the learnable query $\mathbf{Q}_q$ at $\sigma{=}1$ and recover the clean latent and decoded field in one forward pass:
\begin{equation}
    \hat{\mathbf{Z}}_q \!=\! \mathbf{Q}_q - v_\theta\big(\mathbf{Q}_q,\sigma{=}1\,\big|\,\mathbf{U},\mathbf{B}_q,\mathbf{C}_q\big),\quad \quad
    \hat{\mathbf{Y}}_q = \mathcal{D}(\hat{\mathbf{Z}}_q).
    \label{eq:readout}
\end{equation}
We reuse the pretrained velocity head and timestep modulation to adapt generative priors into a fast deterministic solver, where $\mathcal{D}$ is the VAE decoder.

\subsection{Local Query Conditions for RSSM and SSD}
\label{sec:conditions}

RSSM and SSD rely on different geometric manifolds: a horizontal plane at height $z_q$ vs.\ a Tx/Rx-centered sphere. Rather than forcing the generator to learn an implicit planar-to-spherical projection from $\mathbf{I}_{\mathrm{bev}}$ alone, we render 3D urban geometry directly on each target's native manifold to construct $\mathcal{G}_q$ (all token interfaces are summarized in Tab.~\ref{tab:query_slots}).

\paragraph{Planar Query for Multi-Height RSSM ($\mathbf{R}_{z_q}$).}
At query altitude $z_q$, signal coverage depends on building intersections and relative roof heights. Thus, we form a two-channel planar condition over the horizontal grid $(x,y)$:
\begin{equation}
    \mathcal{G}_{\mathrm{RSSM}}(x,y)=\big[\,\mathbb{1}[\mathbf{H}(x,y)\ge z_q],\ \mathbf{H}(x,y)-z_q\,\big],
    \label{eq:planar}
\end{equation}
where $\mathbb{1}[\cdot]$ is an indicator function and $\mathbf{H}(x,y)-z_q$ captures relative building elevation.

\paragraph{Spherical Query for Bidirectional SSD ($\mathbf{S}^{\mathrm{Rx}}, \mathbf{S}^{\mathrm{Tx}}$).}
From the queried location ($\mathbf{r}$ or $\mathbf{t}$), we cast rays along each $1^\circ$ grid direction $\boldsymbol{\omega}_{m,n}$ on the $90\times 360$ hemisphere to construct a three-channel spherical panorama condition:
\begin{equation}
    \mathcal{G}_{\mathrm{SSD}}(m,n)\!=\!\Big[\,h_{m,n},\ \frac{h_{m,n}}{1+d_{m,n}/d_0},\ \alpha\, e^{\frac{-\Delta_{m,n}^2}{2\sigma_\omega^2}}\Big],
    \label{eq:pano}
\end{equation}
which records building-hit occupancy $h_{m,n}\in\{0,1\}$, building-hit distance $d_{m,n}$ normalized by a proximity scale $d_0{=}20\,\mathrm{m}$, and a directional Gaussian prior centered on the opposite endpoint with angular offset $\Delta_{m,n}$, angular width $\sigma_\omega{=}2^\circ$, and LoS/NLoS visibility weight $\alpha$ ($\alpha{=}1$ if unobstructed and $\alpha{=}0.5$ if blocked).

\begin{table}[t]
  \centering
  \small
  \setlength{\tabcolsep}{6pt}
  \renewcommand{\arraystretch}{1.08}
  \caption{\textbf{Instantiation of the unified RadioWorld token interface.} RF radiance field queries (RSSM and bidirectional SSD) and backbone-enabled extensions.}
  \label{tab:query_slots}
  \begin{tabular}{@{}lllll@{}}
    \toprule
    \textbf{Query / Mode} & $\mathbf{I}_{\mathrm{bev}}$ \textbf{(Rx)} & $\mathbf{B}_q$ \textbf{(conditions)} & $\mathbf{C}_q$ \textbf{(9-D input)} & \textbf{Output size} \\
    \midrule
    RSSM ($\mathbf{R}_{z_q}$) & Zeroed & Planar slice & $[\mathbf{t}, z_q, \mathbf{0}]$ & $128{\times}128$ \\
    SSD ($\mathbf{S}^{\mathrm{Rx}} / \mathbf{S}^{\mathrm{Tx}}$) & Retained & Panorama & $[\mathbf{t}, \mathbf{r}, \mathbf{t}-\mathbf{r}]$ & $90{\times}360$ \\
    \midrule
    \multicolumn{5}{@{}l}{\textit{World-Model Extensions (Sec.~\ref{sec:extensions})}} \\
    Multi-frame SSD & Rx trajectory & $T{\times}$ Panorama & $T{\times}[\mathbf{t},\mathbf{r}_{1:T},\mathbf{t}-\mathbf{r}_{1:T}]$ & $T{\times}90{\times}360$ \\
    Text Reasoning & Retained & \textit{Off} & \textit{Off} & Text tokens \\
    \bottomrule
  \end{tabular}
\end{table}

\subsection{World Model-Enabled Functional Extensions}
\label{sec:extensions}

Because RadioWorld preserves both the temporal generation capacity of its video world-model backbone and the autoregressive language interface of its VLM understanding stream, the core architecture naturally supports two extensions without structural redesign.

\paragraph{Multi-frame SSD Synthesis (Spatiotemporal Extension).}
To predict SSDs along a mobile receiver path in a single forward pass, we draw the trajectory onto the Rx channels of $\mathbf{I}_{\mathrm{bev}}$ and synthesize all frames jointly. Cross-frame attention follows spatial geometry rather than frame order.

\paragraph{Link-Level Propagation Reasoning (Language Extension).}
With the field-generation stream switched off, the Scene-Understanding Transformer operates directly as a causal VLM. Appending a text prompt $\mathbf{P}$ (specifying $\mathbf{t}$, $\mathbf{r}$, and the six BEV channel semantics) after encoded BEV tokens $\mathbf{U}^{(0)}$ enables autoregressive decoding of link-level propagation descriptions, such as identifying the dominant AoA/AoD of multi-path propagation.

\subsection{Training Objective and Inference}
\label{sec:training}

All field-synthesis queries share a dual-domain regression loss over both the VAE latent and the decoded field:
\begin{equation}
  \mathcal{L}_{\mathrm{field}}(\hat{\mathbf{Y}},\mathbf{Y})
  =\lambda_z \|\hat{\mathbf{Z}}\!-\!\mathbf{Z}\|_2^2
  +\lambda_y \|\hat{\mathbf{Y}}\!-\!\mathbf{Y}\|_2^2,
  \label{eq:loss}
\end{equation}
where $\mathbf{Z}=\mathcal{E}(\mathbf{Y})$ denotes the VAE-encoded latent of the ground-truth field $\mathbf{Y}$, and $(\hat{\mathbf{Z}}, \hat{\mathbf{Y}})$ are the single-step predictions with $\hat{\mathbf{Y}}=\mathcal{D}(\hat{\mathbf{Z}})$ from Eq.~\eqref{eq:readout}.
This yields $\mathcal{L}_{\mathrm{RSSM}} = \mathcal{L}_{\mathrm{field}}(\hat{\mathbf{R}}_{z_q},\mathbf{R}_{z_q})$, $\mathcal{L}_{\mathrm{SSD}} = \mathcal{L}_{\mathrm{field}}(\hat{\mathbf{S}},\mathbf{S})$ with $\mathbf{S}\in\{\mathbf{S}^{\mathrm{Rx}},\mathbf{S}^{\mathrm{Tx}}\}$, and $\mathcal{L}_{\mathrm{traj}} = \mathcal{L}_{\mathrm{field}}(\hat{\mathbf{S}}_{1:T},\mathbf{S}_{1:T})$ averaged over all $T{=}64$ trajectory frames. The text reasoning is trained with next-token cross-entropy over the $M$-token ground-truth propagation description $(y_1,\ldots,y_M)$, conditioned on encoded BEV tokens $\mathbf{U}^{(0)}$ and text prompt $\mathbf{P}$:
\begin{equation}
    \mathcal{L}_{\mathrm{text}}=-\sum_{t=1}^{M}\log p_\theta(y_t\mid \mathbf{U}^{(0)}, \mathbf{P}, y_{<t}),
    \label{eq:loss-text}
\end{equation}
where $y_{<t}=(y_1,\ldots,y_{t-1})$ denotes the preceding answer tokens.
During joint training, optimization steps cycle in equal proportion across RSSM, SSD (alternating AoA and AoD), multi-frame SSD, and link-level propagation reasoning. At inference, each RSSM, single-location SSD, or $64$-frame trajectory SSD is synthesized in a single feed-forward pass purely from 3D scene geometry and transceiver coordinates, without any target-scene RF measurements.

\section{UrbanRF-Bench}
\label{sec:benchmark}

\begin{table}[t]
  \centering
  \small
  \setlength{\tabcolsep}{8pt}
  \renewcommand{\arraystretch}{1.05}
  \caption{\textbf{UrbanRF-Bench structure.} Test cities (Beijing, Washington) are disjoint from training/validation cities.}
  \label{tab:urbanrf}
  \begin{tabular}{@{}lrrr@{}}
    \toprule
    & \textbf{Train} & \textbf{Validation} & \textbf{Test} \\
    \midrule
    Cities / scenes & 5 / 419 & 5 / 47 & 2 / 191 \\
    RSSMs ($1.5, 10, 20, 30\,\mathrm{m}$) & 83,800 & 9,400 & 38,200 \\
    Static SSD pairs (Rx and Tx) & 1,746,765 & 187,789 & 561,896 \\
    Multi-frame SSDs ($64$-frame) & 25,005 & 2,259 & 5,330 \\
    RF propagation descriptions & 1,746,765 & 187,789 & 561,896 \\
    \bottomrule
  \end{tabular}
\end{table}

Existing radio datasets treat spatial and angular modeling in isolation, providing either RSSMs~\cite{levie2021radiounet,yapar2023challenge} or SSDs limited to small indoor scenes~\cite{zhao2023nerf2,wen2025wrfgs}. To unify both modalities for cross-city RF radiance field synthesis, we construct \textbf{UrbanRF-Bench} with Sionna RT~\cite{ait_aoudia2025sionnart}.

\begin{figure}[t]
  \centering
  \includegraphics[width=0.92\textwidth]{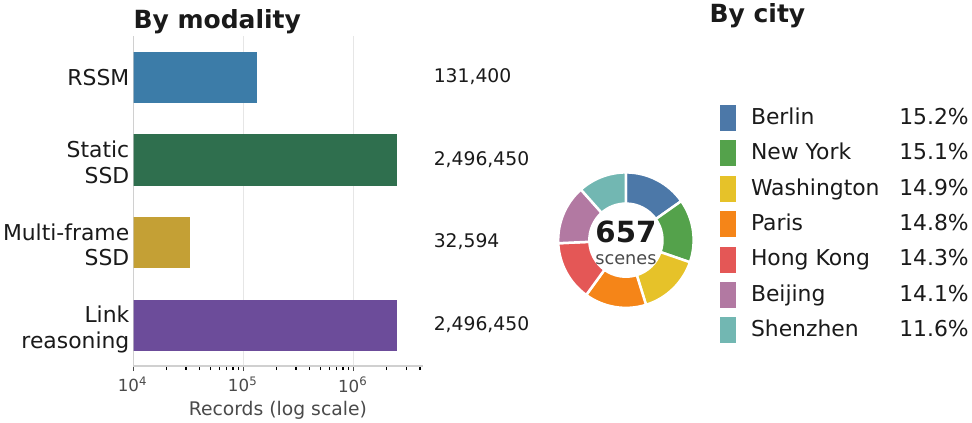}
  \caption{\textbf{Composition of UrbanRF-Bench.} Left: record counts per modality on a log scale. Static SSD counts a pair of Rx and Tx SSDs. Right: the $657$ scenes across seven cities.}
  \label{fig:suppl_urbanrf_dist}
\end{figure}

\subsection{Dataset Composition and Cross-City Split}
UrbanRF-Bench spans $657$ urban tiles ($128\times128\,\mathrm{m}$ at $1\,\mathrm{m}$ resolution) extracted from OpenStreetMap across seven global cities covering four distinct architectural layouts: dense high-rise districts (Hong Kong, Shenzhen), regular grid layouts (New York, Beijing), courtyard blocks (Berlin, Paris), and mid-rise residential areas (Washington). As illustrated in Fig.~\ref{fig:suppl_urbanrf_dist} and summarized in Tab.~\ref{tab:urbanrf}, we enforce a strict city-level split: five cities form the training ($419$ scenes) and validation ($47$ scenes) splits, including Berlin ($90$/$10$), Hong Kong ($85$/$9$), New York ($89$/$10$), Paris ($87$/$10$), and Shenzhen ($68$/$8$), while two completely unseen cities, Beijing ($93$ scenes) and Washington ($98$ scenes), constitute the $191$-scene zero-shot test split.

Across all $657$ scenes with $50$ transmitters per scene ($32{,}850$ RF radiance fields in total), UrbanRF-Bench provides $131{,}400$ multi-height RSSMs ($83{,}800$ train, $9{,}400$ val, $38{,}200$ test), $2{,}496{,}450$ static bidirectional SSDs (each storing both Rx- and Tx-side spectra), and $32{,}594$ multi-frame SSD trajectories ($25{,}005$ train, $2{,}259$ val, $5{,}330$ test) of $64$ frames each.

For link-level propagation reasoning, the azimuth domain $[0^\circ, 360^\circ)$ is partitioned into twelve non-overlapping $30^\circ$ sectors. Multipath powers are aggregated per sector separately for arrival (Rx) and departure (Tx) azimuths. The dominant sector is verbalized as ``arrives mainly from'' (Rx) and ``leaves mainly toward'' (Tx). In total, the reasoning dataset covers $2{,}496{,}450$ links, comprising $1{,}746{,}765$ training descriptions alongside the full validation ($187{,}789$) and test ($561{,}896$) splits.

\subsection{Sionna RT Simulation}
All scenes are simulated in Sionna RT~\cite{ait_aoudia2025sionnart} at $3.5\,\mathrm{GHz}$ ($100\,\mathrm{MHz}$ bandwidth, $30\,\mathrm{dBm}$ transmit power) using ITU brick buildings and concrete ground, enabling LoS, specular and diffuse reflection, refraction, and diffraction up to $5$ interactions. In each tile, $50$ rooftop-mounted transmitters are placed $5$--$10\,\mathrm{m}$ above building roofs. For each transmitter, we sample RSSMs at four altitudes ($\{1.5, 10, 20, 30\}\,\mathrm{m}$), up to $25$ static receivers per height, and $64$-frame receiver trajectories at $0.5\,\mathrm{m}$ spacing ($\sim 31.5\,\mathrm{m}$ path length). Ground-truth bidirectional SSDs are extracted via Bartlett beamforming~\cite{vantrees2002optimum,zhao2023nerf2} using virtual $4\times 4$ half-wavelength planar arrays. Finally, all RSSMs and SSDs are converted to the dB scale and min--max normalized to $[0,1]$.

\section{Experiments}
\label{sec:experiments}

\subsection{Experimental Setup}

\paragraph{Protocols and Metrics.}
For \textbf{RSSM}, we compare against RadioUNet~\cite{levie2021radiounet}, RadioDiff~\cite{wang2025radiodiff}, and RadioMamba~\cite{jia2025radiomamba}, all retrained from scratch on the five-city training split across all four heights and evaluated on all $191$ unseen test scenes ($38{,}200$ maps).
For \textbf{SSD}, measurement-based baselines require target-scene RF measurements. Thus, we follow their protocol to randomly select one held-out scene from Beijing and one from Washington and report the averaged metrics across both scenes (with the Washington scene visualized in Fig.~\ref{fig:pas_visual}). In each held-out scene, $500$ reference SSDs inside a $40\,\mathrm{m}$-diameter circle are given to NeRF$^2$~\cite{zhao2023nerf2}, GSRF~\cite{yang2025gsrf}, and WRF-GS+~\cite{wen2025wrfgsplus} (optimized per scene), and to GRaF~\cite{yang2026graf} (pretrained across cities). RadioWorld uses zero reference SSDs. We report NMSE, RMSE, PSNR, and SSIM~\cite{wang2004ssim} for both RSSM and SSD.

\paragraph{Implementation Details.}
RadioWorld is initialized from Cosmos 3~\cite{nvidia2026cosmos3} with SigLIP~2, the patch merger, and the Wan~2.2 VAE frozen. We train the adapters, coordinate/task embeddings, query latents, generator projections, language head, timestep embedder, MLPs, norms, and rank-$64$ LoRA~\cite{hu2022lora} ($1.38$B trainable parameters in total) using AdamW~\cite{loshchilov2019adamw} ($\lambda_z=\lambda_y=1$, batch size $32$). Training takes approximately 12 hours on two RTX Pro 6000 GPUs.

\subsection{Main Results}

\begin{figure}[!ht]
  \centering
  \includegraphics[width=0.66\textwidth]{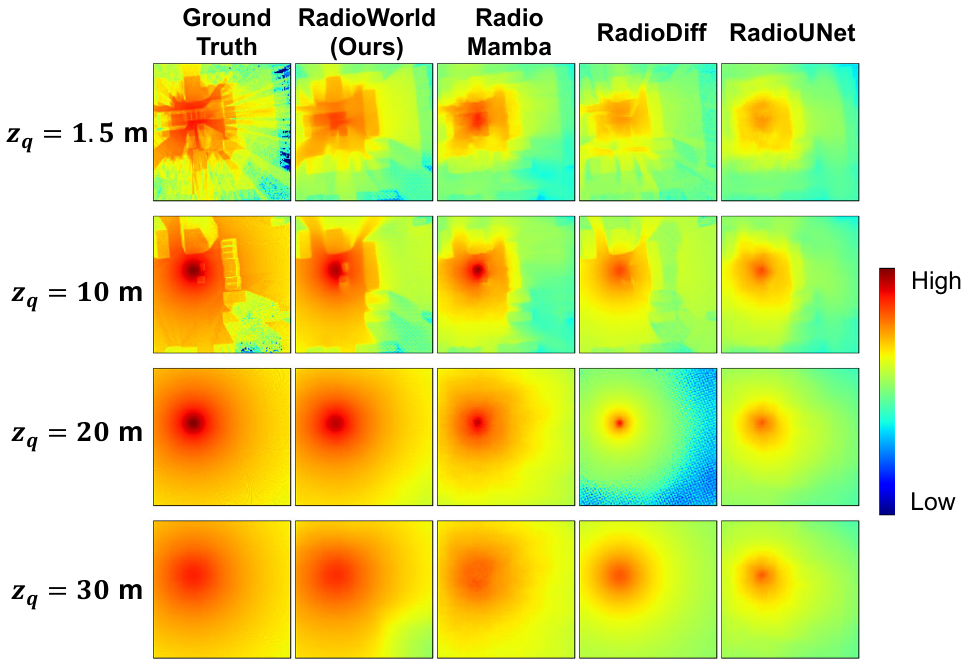}
  \caption{\textbf{Visual comparison of zero-shot multi-height RSSM synthesis ($1.5\,\mathrm{m}$--$30\,\mathrm{m}$) in unseen cities.} RadioWorld closely matches the ground truth in both sharp street-level building shadows ($1.5\,\mathrm{m}$, $10\,\mathrm{m}$) and smooth free-space propagation at higher altitudes ($20\,\mathrm{m}$, $30\,\mathrm{m}$), whereas baselines lose fine details around building edges or introduce severe artifacts.}
  \label{fig:pcm_visual}
\end{figure}

\paragraph{RSSM Synthesis.}
Tab.~\ref{tab:pcm} and Fig.~\ref{fig:pcm_visual} compare zero-shot RSSM synthesis in the two unseen test cities. Across all $38{,}200$ maps, RadioWorld outperforms every specialized radio-map baseline in NMSE and PSNR at every height:
(i) At street-level planes ($1.5\,\mathrm{m}$ and $10\,\mathrm{m}$), radio propagation is largely affected by urban structures. Convolutional and state-space baselines (RadioUNet and RadioMamba) tend to over-smooth these shadow-zone boundaries (Fig.~\ref{fig:pcm_visual}), whereas RadioWorld leverages global propagation features from the Scene-Understanding Transformer to preserve the edge details of the shadow zone.
(ii) The $20\,\mathrm{m}$ height plane intersects typical urban rooftops, creating edge diffraction that makes this height the most challenging for all models. Notably, RadioDiff suffers severe performance degradation, as the multi-step sampling procedure produces less accurate shadow-zone details. In contrast, RadioWorld reduces $20\,\mathrm{m}$ NMSE by $61\%$ over RadioDiff and $38\%$ over RadioMamba while achieving a comparable SSIM to the best baseline. (iii) At $30\,\mathrm{m}$, RF propagation transitions toward line-of-sight (LoS). RadioWorld achieves $31.25\,\mathrm{dB}$ PSNR ($+4.55\,\mathrm{dB}$ and $-57\%$ NMSE over RadioMamba), confirming consistent cross-height generalization in unseen cities.

\begin{table}[!ht]
  \centering
  \footnotesize
  \setlength{\tabcolsep}{8pt}
  \renewcommand{\arraystretch}{0.96}
  \caption{\textbf{Zero-shot RSSM prediction in unseen cities} ($38{,}200$ maps), per query height.}
  \label{tab:pcm}
  \begin{tabular}{@{}llcccc@{}}
    \toprule
    \textbf{Method} & \textbf{Height} & NMSE$\downarrow$ & RMSE$\downarrow$ & PSNR$\uparrow$ & SSIM$\uparrow$ \\
    \midrule
    \multirow{5}{*}{RadioUNet~\cite{levie2021radiounet}}
      & $1.5\,\mathrm{m}$ & 0.0477 & 0.1102 & 20.10 & 0.6976 \\
      & $10\,\mathrm{m}$  & 0.0385 & 0.1006 & 21.06 & 0.7126 \\
      & $20\,\mathrm{m}$  & 0.0556 & 0.1149 & 20.06 & 0.6667 \\
      & $30\,\mathrm{m}$  & 0.0285 & 0.0800 & 23.91 & 0.9380 \\
      & All               & 0.0426 & 0.1014 & 21.28 & 0.7537 \\
    \midrule
    \multirow{5}{*}{RadioDiff~\cite{wang2025radiodiff}}
      & $1.5\,\mathrm{m}$ & 0.0369 & 0.0992 & 21.15 & 0.6916 \\
      & $10\,\mathrm{m}$  & 0.0336 & 0.0958 & 21.58 & 0.7026 \\
      & $20\,\mathrm{m}$  & 0.0615 & 0.1383 & 18.29 & 0.5246 \\
      & $30\,\mathrm{m}$  & 0.0201 & 0.0686 & 25.68 & 0.9373 \\
      & All               & 0.0380 & 0.1005 & 21.68 & 0.7140 \\
    \midrule
    \multirow{5}{*}{RadioMamba~\cite{jia2025radiomamba}}
      & $1.5\,\mathrm{m}$ & 0.0361 & 0.0966 & 21.38 & 0.7193 \\
      & $10\,\mathrm{m}$  & 0.0285 & 0.0867 & 22.57 & 0.7337 \\
      & $20\,\mathrm{m}$  & 0.0392 & 0.1002 & 21.36 & \textbf{0.6680} \\
      & $30\,\mathrm{m}$  & 0.0189 & 0.0633 & 26.70 & 0.9416 \\
      & All               & 0.0307 & 0.0867 & 23.00 & 0.7657 \\
    \midrule
    \cellcolor{metablue!12}
      & \cellcolor{metablue!12}$1.5\,\mathrm{m}$ & \cellcolor{metablue!12}\textbf{0.0250} & \cellcolor{metablue!12}\textbf{0.0785} & \cellcolor{metablue!12}\textbf{23.35} & \cellcolor{metablue!12}\textbf{0.7259} \\
    \cellcolor{metablue!12}
      & \cellcolor{metablue!12}$10\,\mathrm{m}$  & \cellcolor{metablue!12}\textbf{0.0159} & \cellcolor{metablue!12}\textbf{0.0641} & \cellcolor{metablue!12}\textbf{25.14} & \cellcolor{metablue!12}\textbf{0.7459} \\
    \cellcolor{metablue!12}
      & \cellcolor{metablue!12}$20\,\mathrm{m}$  & \cellcolor{metablue!12}\textbf{0.0242} & \cellcolor{metablue!12}\textbf{0.0775} & \cellcolor{metablue!12}\textbf{23.85} & \cellcolor{metablue!12}0.6661 \\
    \cellcolor{metablue!12}
      & \cellcolor{metablue!12}$30\,\mathrm{m}$  & \cellcolor{metablue!12}\textbf{0.0081} & \cellcolor{metablue!12}\textbf{0.0368} & \cellcolor{metablue!12}\textbf{31.25} & \cellcolor{metablue!12}\textbf{0.9471} \\
    \cellcolor{metablue!12}\multirow{-5}{*}{\textbf{RadioWorld}}
      & \cellcolor{metablue!12}All               & \cellcolor{metablue!12}\textbf{0.0183} & \cellcolor{metablue!12}\textbf{0.0642} & \cellcolor{metablue!12}\textbf{25.90} & \cellcolor{metablue!12}\textbf{0.7712} \\
    \bottomrule
  \end{tabular}
\end{table}

\begin{figure}[!t]
  \centering
  \includegraphics[width=0.90\textwidth]{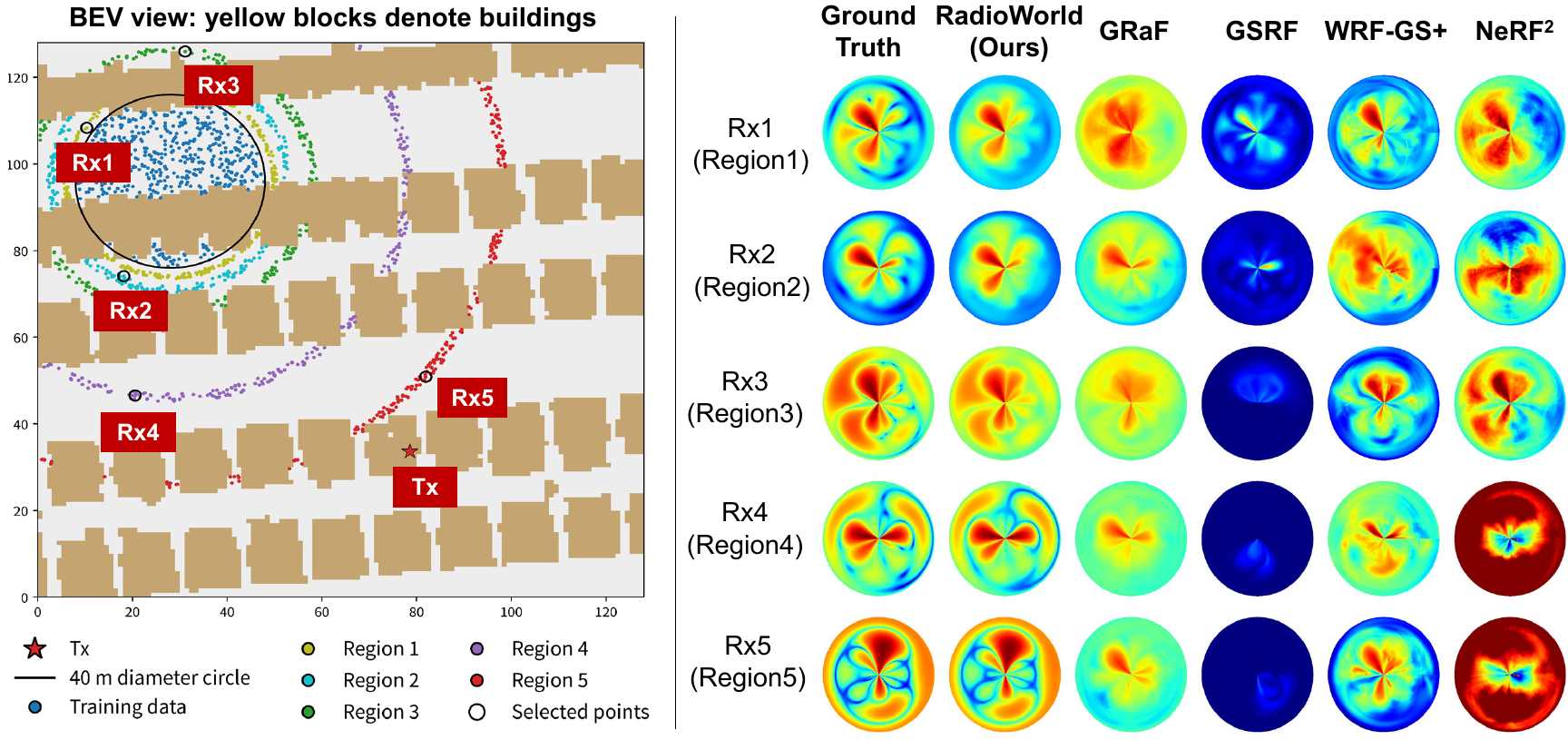}
  \caption{\textbf{Visual comparison of SSD synthesis.} \textbf{Left:} Tx (red star), reference region (black circle) with $500$ measurements given to baselines (blue), and test receivers $\mathrm{Rx1}$--$\mathrm{Rx5}$ at $2$, $5$, $10$, $30$, and $50\,\mathrm{m}$ outside the reference region. \textbf{Right:} measurement-based baselines lose the dominant lobes, whereas RadioWorld synthesizes high-fidelity SSD at all regions without any reference measurements.}
  \label{fig:pas_visual}
\end{figure}

\paragraph{SSD Synthesis.}
Tab.~\ref{tab:pas} and Fig.~\ref{fig:pas_visual} compare RadioWorld against measurement-based neural RF field methods. As test queries move outward from Region 1 to Region 5, WRF-GS+ NMSE worsens by $2.65\times$ from $0.0685$ to $0.1815$, and GRaF SSIM drops by $46.5\%$ from $0.7143$ to $0.3820$. From Fig.~\ref{fig:pas_visual}, measurement-based models lose the dominant lobes outside the measurement region because they primarily learn the local RF radiance field within the reference measurement region. In contrast, RadioWorld uses \emph{zero} reference SSD measurements and predicts SSD purely from scene geometry and transceiver coordinates. Consequently, it achieves consistent predictive performance across all query distances.

\begin{table}[!ht]
  \centering
  \scriptsize
  \setlength{\tabcolsep}{7.5pt}
  \renewcommand{\arraystretch}{0.92}
  \caption{\textbf{SSD synthesis in unseen cities} (averaged over one held-out scene from Beijing and one from Washington; Fig.~\ref{fig:pas_visual} illustrates the Washington scene). Regions 1–5 correspond to query locations at distances of 2, 5, 10, 30, and 50 m beyond the boundary of the 40 m-diameter reference circle, respectively. Per-scene and cross-scene baselines use $500$ reference SSDs per scene. RadioWorld uses zero reference SSDs. ``All'' aggregates Regions 1--5.}
  \label{tab:pas}
  \begin{tabular}{@{}llcccc@{}}
    \toprule
    \textbf{Method} & \textbf{Region} & NMSE$\downarrow$ & RMSE$\downarrow$ & PSNR$\uparrow$ & SSIM$\uparrow$ \\
    \midrule
    \multicolumn{6}{@{}l}{\textit{Per-scene, with reference SSD measurements}} \\
    \multirow{6}{*}{NeRF$^2$~\cite{zhao2023nerf2}}
      & Region 1 ($2\,\mathrm{m}$)  & 0.1505 & 0.1904 & 15.50 & 0.5775 \\
      & Region 2 ($5\,\mathrm{m}$)  & 0.2783 & 0.2522 & 13.01 & 0.4995 \\
      & Region 3 ($10\,\mathrm{m}$) & 0.2510 & 0.2612 & 12.38 & 0.4343 \\
      & Region 4 ($30\,\mathrm{m}$) & 0.8829 & 0.5378 & 5.69  & 0.2138 \\
      & Region 5 ($50\,\mathrm{m}$) & 0.7078 & 0.4640 & 7.05  & 0.2347 \\
      & All                         & 0.4554 & 0.3416 & 10.72 & 0.3917 \\
    \midrule
    \multirow{6}{*}{WRF-GS+~\cite{wen2025wrfgsplus}}
      & Region 1 ($2\,\mathrm{m}$)  & 0.0685 & 0.1281 & \textbf{19.35} & 0.6756 \\
      & Region 2 ($5\,\mathrm{m}$)  & 0.0894 & 0.1570 & 16.69 & 0.6058 \\
      & Region 3 ($10\,\mathrm{m}$) & 0.1106 & 0.1909 & 14.74 & 0.5165 \\
      & Region 4 ($30\,\mathrm{m}$) & 0.1345 & 0.2158 & 13.43 & 0.3606 \\
      & Region 5 ($50\,\mathrm{m}$) & 0.1815 & 0.2487 & 12.23 & 0.3278 \\
      & All                         & 0.1169 & 0.1881 & 15.29 & 0.4971 \\
    \midrule
    \multirow{6}{*}{GSRF~\cite{yang2025gsrf}}
      & Region 1 ($2\,\mathrm{m}$)  & 0.3449 & 0.3210 & 10.12 & 0.3715 \\
      & Region 2 ($5\,\mathrm{m}$)  & 0.6736 & 0.4601 & 6.81  & 0.1882 \\
      & Region 3 ($10\,\mathrm{m}$) & 0.8395 & 0.5376 & 5.47  & 0.0943 \\
      & Region 4 ($30\,\mathrm{m}$) & 0.9334 & 0.5769 & 4.80  & 0.0453 \\
      & Region 5 ($50\,\mathrm{m}$) & 0.9409 & 0.5725 & 4.87  & 0.0395 \\
      & All                         & 0.7459 & 0.4933 & 6.42  & 0.1481 \\
    \midrule
    \multicolumn{6}{@{}l}{\textit{Cross-scene, with reference SSD measurements}} \\
    \multirow{6}{*}{GRaF~\cite{yang2026graf}}
      & Region 1 ($2\,\mathrm{m}$)  & 0.0769 & 0.1428 & 17.53 & \textbf{0.7143} \\
      & Region 2 ($5\,\mathrm{m}$)  & 0.0850 & 0.1537 & 16.69 & 0.6782 \\
      & Region 3 ($10\,\mathrm{m}$) & 0.0975 & 0.1765 & 15.33 & 0.6062 \\
      & Region 4 ($30\,\mathrm{m}$) & 0.1125 & 0.1987 & 14.14 & 0.4464 \\
      & Region 5 ($50\,\mathrm{m}$) & 0.1603 & 0.2341 & 12.75 & 0.3820 \\
      & All                         & 0.1065 & 0.1812 & 15.29 & 0.5650 \\
    \midrule
    \multicolumn{6}{@{}l}{\textit{Cross-scene, without reference SSD measurements}} \\
    \cellcolor{metablue!12}
      & \cellcolor{metablue!12}Region 1 ($2\,\mathrm{m}$)  & \cellcolor{metablue!12}\textbf{0.0574} & \cellcolor{metablue!12}\textbf{0.1278} & \cellcolor{metablue!12}18.27 & \cellcolor{metablue!12}0.7122 \\
    \cellcolor{metablue!12}
      & \cellcolor{metablue!12}Region 2 ($5\,\mathrm{m}$)  & \cellcolor{metablue!12}\textbf{0.0585} & \cellcolor{metablue!12}\textbf{0.1301} & \cellcolor{metablue!12}\textbf{18.08} & \cellcolor{metablue!12}\textbf{0.7035} \\
    \cellcolor{metablue!12}
      & \cellcolor{metablue!12}Region 3 ($10\,\mathrm{m}$) & \cellcolor{metablue!12}\textbf{0.0529} & \cellcolor{metablue!12}\textbf{0.1280} & \cellcolor{metablue!12}\textbf{18.51} & \cellcolor{metablue!12}\textbf{0.7372} \\
    \cellcolor{metablue!12}
      & \cellcolor{metablue!12}Region 4 ($30\,\mathrm{m}$) & \cellcolor{metablue!12}\textbf{0.0330} & \cellcolor{metablue!12}\textbf{0.1000} & \cellcolor{metablue!12}\textbf{20.83} & \cellcolor{metablue!12}\textbf{0.8145} \\
    \cellcolor{metablue!12}
      & \cellcolor{metablue!12}Region 5 ($50\,\mathrm{m}$) & \cellcolor{metablue!12}\textbf{0.0349} & \cellcolor{metablue!12}\textbf{0.0979} & \cellcolor{metablue!12}\textbf{20.99} & \cellcolor{metablue!12}\textbf{0.8133} \\
    \cellcolor{metablue!12}\multirow{-6}{*}{\textbf{RadioWorld}}
      & \cellcolor{metablue!12}All                         & \cellcolor{metablue!12}\textbf{0.0473} & \cellcolor{metablue!12}\textbf{0.1167} & \cellcolor{metablue!12}\textbf{19.34} & \cellcolor{metablue!12}\textbf{0.7563} \\
    \bottomrule
  \end{tabular}
\end{table}

\paragraph{Multi-Frame Trajectory SSD Synthesis.}
To verify that RadioWorld inherits spatiotemporal video generation capabilities from its backbone, we evaluate $64$-frame trajectory SSD synthesis on $5{,}330$ trajectories (Tab.~\ref{tab:pas-video}). Considering the complicated multipath propagation between the Tx and the Rx, the model cannot rely on simple LoS priors and must infer evolving multipath reflections across trajectories. Under this challenging scenario, single-forward multi-frame SSD synthesis matches independent per-frame prediction in fidelity across both Tx-side and Rx-side SSDs, while reducing inference latency by $42\%$ from ${\sim}1.05\,\mathrm{s}$ to ${\sim}0.60\,\mathrm{s}$.

\begin{table}[!ht]
  \centering
  \small
  \setlength{\tabcolsep}{8pt}
  \renewcommand{\arraystretch}{1.05}
  \caption{\textbf{Multi-frame SSD trajectory synthesis on test tracks without reference measurements} in Beijing and Washington ($5{,}330$ trajectories). Single-forward trajectory synthesis matches per-frame prediction while reducing forward latency by $42\%$.}
  \label{tab:pas-video}
  \begin{tabular}{@{}llccccc@{}}
    \toprule
    & & NMSE$\downarrow$ & RMSE$\downarrow$ & PSNR$\uparrow$ & SSIM$\uparrow$ & Time (s) \\
    \midrule
    \multirow{2}{*}{Tx-side}
      & Trajectory-based & 0.0472 & 0.1115 & 19.65 & 0.7611 & \textbf{0.604} \\
      & Per-frame-based  & 0.0472 & 0.1113 & 19.67 & 0.7616 & 1.046 \\
    \midrule
    \multirow{2}{*}{Rx-side}
      & Trajectory-based & 0.0687 & 0.1398 & 17.55 & 0.7156 & \textbf{0.603} \\
      & Per-frame-based  & 0.0690 & 0.1399 & 17.55 & 0.7150 & 1.044 \\
    \bottomrule
  \end{tabular}
\end{table}

\begin{figure}[!h]
  \centering
  \includegraphics[width=0.45\textwidth]{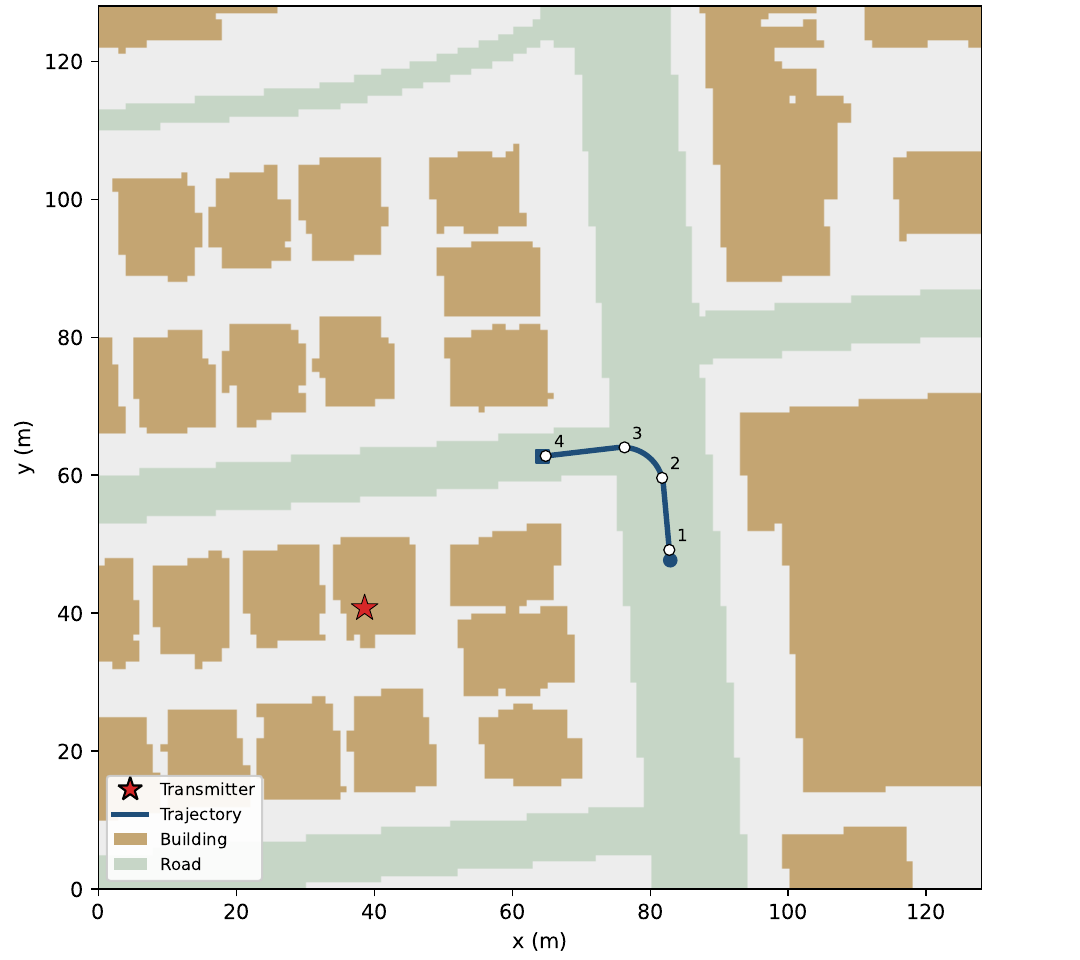}
  \caption{\textbf{A turning trajectory in an unseen Washington scene.}
  The $128\times128\,\mathrm{m}$ tile contains $66$ buildings.
  The star is the transmitter, and the curve is the $64$-frame receiver path.
  Markers $1$--$4$ are frames $3$, $24$, $39$, and $62$.}
  \label{fig:suppl_turn_map}
\end{figure}

\begin{figure}[!t]
  \centering
  \includegraphics[width=0.84\textwidth]{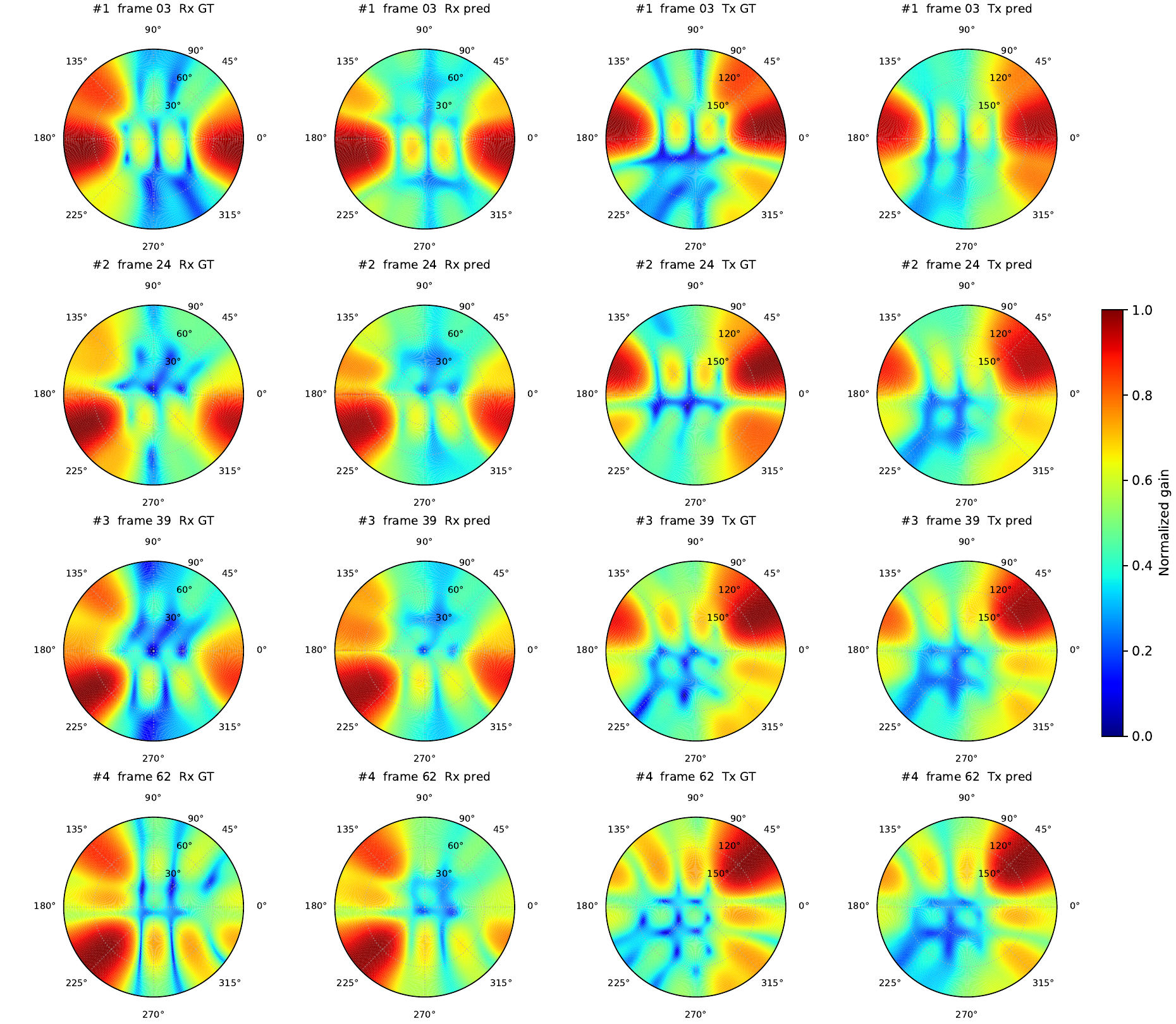}
  \caption{\textbf{Rx-side and Tx-side SSDs along the trajectory in Fig.~\ref{fig:suppl_turn_map}.}
  Each row is one of the four marked frames.
  Columns are Rx ground truth, Rx prediction, Tx ground truth, and Tx prediction.
  Each side's prediction comes from a single forward pass over all $64$ frames.}
  \label{fig:suppl_turn_ssd}
\end{figure}

Fig.~\ref{fig:suppl_turn_map} shows one such $64$-frame mobile Rx trajectory in an unseen Washington scene. The tile is $128\times128\,\mathrm{m}$ and contains $66$ buildings. The Rx moves at a height of $1.5\,\mathrm{m}$ along the trajectory. The Tx is marked by a star, and the four numbered markers are frames $3$, $24$, $39$, and $62$. These frames are spread along the trajectory. Fig.~\ref{fig:suppl_turn_ssd} compares the ground-truth and predicted SSDs at those four positions. Each row is one frame. From left to right, the columns are the Rx-side ground truth SSD, the Rx-side SSD prediction, the Tx-side ground truth SSD, and the Tx-side SSD prediction. Each of the Tx-side and Rx-side multi-frame SSDs is produced by a single forward pass of the pretrained RadioWorld over the full $64$-frame trajectory, using only the scene geometry and the transceiver positions, and the predictions align closely with the ground-truth SSDs.

\paragraph{Link-Level Propagation Reasoning.}
Tab.~\ref{tab:reason} reports the Dominant AoA/AoD Sector Accuracy on RF propagation reasoning (defined over $12$ non-overlapping $30^\circ$ azimuth sectors by aggregating multipath power within each sector). The Scene-Understanding Transformer reads the BEV tokens and identifies the dominant $30^\circ$ AoA and AoD sectors across $561{,}896$ unseen test links with $89.95\%$ AoA and $92.84\%$ AoD accuracy. This confirms that the scene-understanding stream not only acts as an EM propagation feature encoder, but also implicitly internalizes Tx-Rx link-level propagation from scene geometry.

\begin{table}[!ht]
  \centering
  \small
  \setlength{\tabcolsep}{10pt}
  \renewcommand{\arraystretch}{1.05}
  \caption{\textbf{Dominant AoA/AoD Sector Accuracy} ($30^\circ$ azimuth sectors) on $561{,}896$ unseen test links from scene geometry and transceiver coordinates.}
  \label{tab:reason}
  \begin{tabular}{@{}lccc@{}}
    \toprule
    \textbf{Split} & \textbf{Tx--Rx Links} & \textbf{AoA Accuracy} & \textbf{AoD Accuracy} \\
    \midrule
    All        & 561,896 & 89.95\% & 92.84\% \\
    Beijing    & 268,608 & 94.60\% & 95.02\% \\
    Washington & 293,288 & 85.69\% & 90.84\% \\
    \bottomrule
  \end{tabular}
\end{table}

\subsection{Downstream Applications}

\begin{figure}[!ht]
  \centering
  \begin{minipage}[t]{0.48\textwidth}
    \centering
    \includegraphics[width=\linewidth]{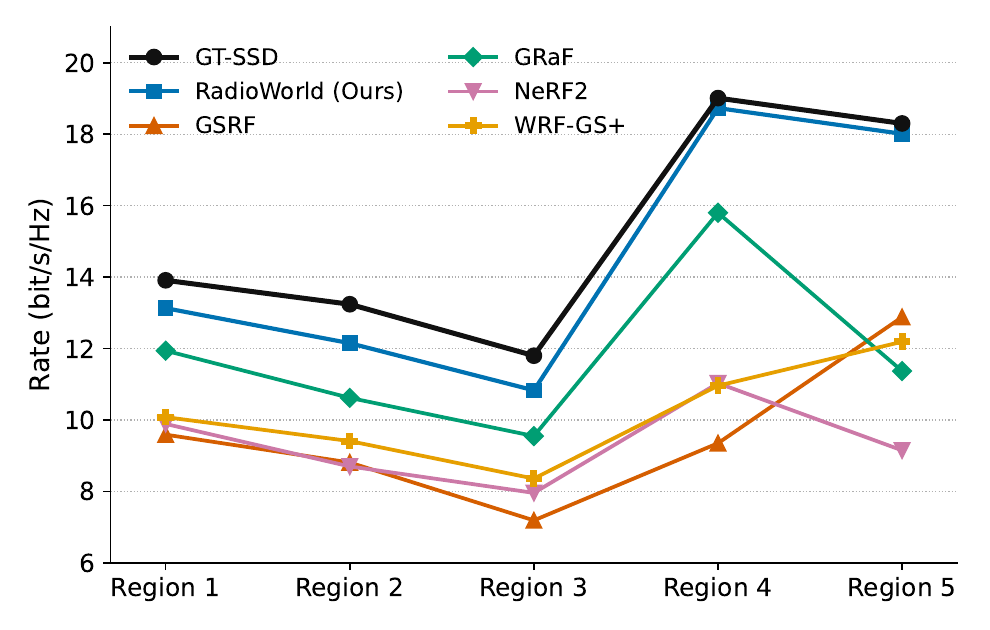}
    \caption{\textbf{Beam selection from predicted SSD} in an unseen Washington scene across Regions 1--5 (see Fig.~\ref{fig:pas_visual}). \emph{GT-SSD} selects optimal beams from the ground-truth SSD.}
    \label{fig:pas_downstream}
  \end{minipage}\hfill
  \begin{minipage}[t]{0.48\textwidth}
    \centering
    \includegraphics[width=\linewidth]{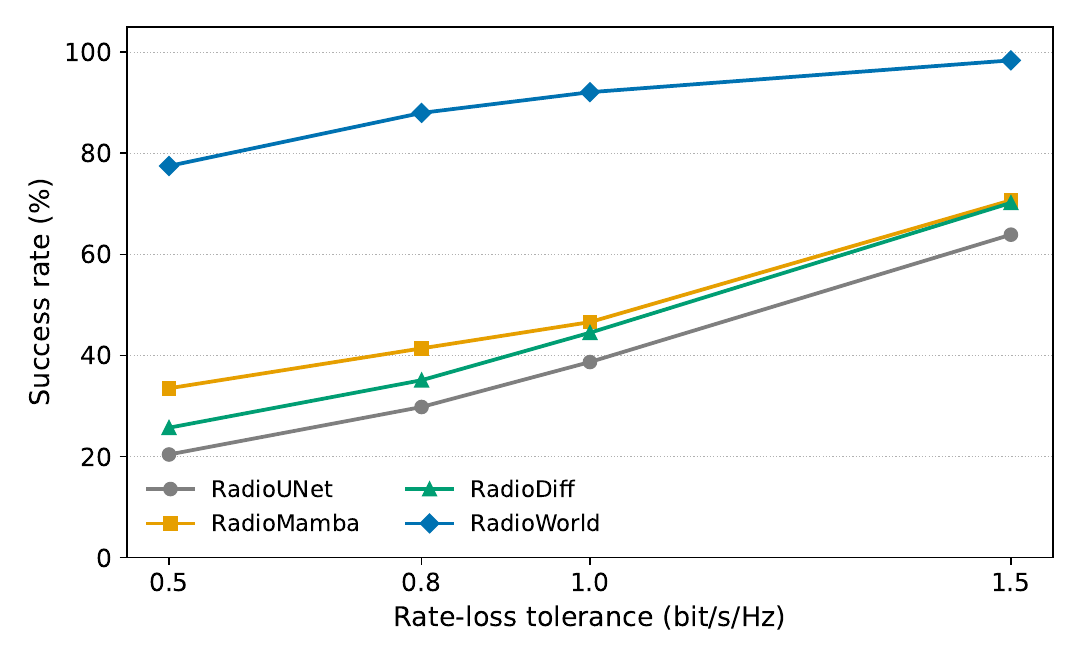}
    \caption{\textbf{Base station (BS) site selection} success rate across $50$ candidate locations under different rate-loss tolerances between the chosen and optimal sites.}
    \label{fig:pcm_downstream}
  \end{minipage}
\end{figure}

\paragraph{Beam Selection from Predicted SSD.}
Beam selection is adopted as a downstream task. Specifically, we consider the scenario in Fig.~\ref{fig:pas_visual} and equip the receiver in each region with an $8\times 8$ half-wavelength planar array and a $64$-beam 2D DFT codebook~\cite{heath2016overview}.
As the peak of the Rx-side SSD implies the AoA of the dominant path, we can choose the codeword aligned with this peak direction. For experiments, we evaluate achievable spectral efficiency $\log_2(1+|\mathbf{w}^{H}\mathbf{h}|^2/N_0)$ on the true ray-traced multipath channel $\mathbf{h}$, where $\mathbf{w}$ denotes the selected beam. Over Regions 1--5 (Fig.~\ref{fig:pas_downstream}), RadioWorld outperforms all measurement-based baselines across every outer ring, closely approaching the GT-SSD upper bound. As query locations move farther away from the reference measurement region, baseline methods suffer performance degradation, while RadioWorld remains aligned with the GT-SSD scheme.

\paragraph{Base Station Site Selection from Predicted RSSM.}
We evaluate whether predicted RSSMs can guide practical base station (BS) site selection across all $191$ unseen test scenes in Beijing and Washington (Fig.~\ref{fig:pcm_downstream}). In each scene, the ground-truth optimal site is the candidate among the $50$ transmitters with the highest four-height mean rate under Sionna RT. Each method selects the candidate with the highest four-height mean rate from its own predicted RSSMs. A selection is successful when the ground-truth Sionna rate of the chosen site lies within a rate-loss tolerance $\delta \in \{0.5, 0.8, 1.0, 1.5\}\,\mathrm{bit/s/Hz}$ of the optimal site. RadioWorld outperforms all baselines at every tolerance. At $\delta = 1.0\,\mathrm{bit/s/Hz}$, it reaches a $92.1\%$ success rate, compared with RadioMamba ($46.6\%$), RadioDiff ($44.5\%$), and RadioUNet ($38.7\%$).

\subsection{Ablation Studies}

\begin{table}[!h]
  \centering
  \small
  \setlength{\tabcolsep}{6pt}
  \renewcommand{\arraystretch}{1.05}
  \caption{\textbf{Design ablations on all unseen test cities} ($38{,}200$ RSSMs and $561{,}896$ receiver SSDs).}
  \label{tab:ablation}
  \begin{tabular}{@{}lccccc@{}}
    \toprule
    \textbf{Variant} & NMSE$\downarrow$ & RMSE$\downarrow$ & PSNR$\uparrow$ & SSIM$\uparrow$ & Time (ms)$\downarrow$ \\
    \midrule
    \multicolumn{6}{@{}l}{\textit{(a) Readout: 8-step flow matching vs.\ single-step ($\sigma{=}1$)}} \\
    RSSM / Flow matching & 0.0396 & 0.0944 & 22.24 & 0.7388 & 158 \\
    RSSM / Single-step & \textbf{0.0183} & \textbf{0.0642} & \textbf{25.90} & \textbf{0.7712} & \textbf{69} \\
    SSD / Flow matching  & 0.0319 & 0.0915 & 22.18 & 0.7970 & 119 \\
    SSD / Single-step & \textbf{0.0184} & \textbf{0.0731} & \textbf{23.79} & \textbf{0.8811} & \textbf{47} \\
    \midrule
    \multicolumn{6}{@{}l}{\textit{(b) SSD conditioning: BEV heatmap vs.\ spherical panorama}} \\
    BEV heatmap & 0.0318 & 0.1094 & 20.09 & 0.8101 & -- \\
    Spherical panorama & \textbf{0.0184} & \textbf{0.0731} & \textbf{23.79} & \textbf{0.8811} & -- \\
    \bottomrule
  \end{tabular}
\end{table}

\paragraph{Single-Step Deterministic Regression.}
We first verify the effectiveness of the single-step deterministic regression of RSSMs and SSDs by comparing it with multi-step flow matching. Tab.~\ref{tab:ablation}a shows the results of $8$-step Euler flow integration from Gaussian noise against our single-step readout from learned query $\mathbf{Q}_q$ at $\sigma{=}1$ across all test scenes. It is observed that single-step regression achieves lower NMSE, with a $2.3\times$ speedup on RSSMs and $2.5\times$ on SSDs over iterative flow matching. This gain arises because sampling Gaussian noise differs across runs, introducing inevitable variance.

\paragraph{Spherical Panorama Conditions for SSD Synthesis.}
We also validate the benefit of the spherical panorama condition for SSD synthesis in Tab.~\ref{tab:ablation}b, by comparing it against a $2$-channel $128\times 128$ BEV Rx Gaussian heatmap condition (encoding Rx horizontal position and normalized elevation). The spherical panorama condition performs better because the BEV Gaussian heatmap only marks where the Rx is in the top-down plane without encoding the surrounding local 3D geometry. In contrast, the spherical panorama condition provides surrounding building information, aligning local 3D visibility with the target SSD.

\begin{table}[!ht]
  \centering
  \small
  \setlength{\tabcolsep}{3.4pt}
  \renewcommand{\arraystretch}{1.05}
  \caption{\textbf{Pretrained world-model initialization vs.\ training from scratch} across all tasks on the unseen test cities ($38{,}200$ RSSMs, $561{,}896\times 2$ Tx and Rx SSDs, $5{,}330\times 2$ trajectory Tx and Rx SSDs, and $561{,}896$ link-level propagation descriptions).}
  \label{tab:ablate-init}
  \begin{tabular}{@{}lcccccccc@{}}
    \toprule
    & \multicolumn{4}{c}{Random Initialization} & \multicolumn{4}{c}{Pretrained World Model Initialization} \\
    \cmidrule(lr){2-5}\cmidrule(l){6-9}
    \textbf{Task} & NMSE$\downarrow$ & RMSE$\downarrow$ & PSNR$\uparrow$ & SSIM$\uparrow$ & NMSE$\downarrow$ & RMSE$\downarrow$ & PSNR$\uparrow$ & SSIM$\uparrow$ \\
    \midrule
    RSSM & 0.0195 & 0.0678 & 25.30 & 0.7458 & \textbf{0.0183} & \textbf{0.0642} & \textbf{25.90} & \textbf{0.7712} \\
    Static SSD (Rx) & 0.0287 & 0.0998 & 20.38 & 0.7486 & \textbf{0.0184} & \textbf{0.0731} & \textbf{23.79} & \textbf{0.8811} \\
    Static SSD (Tx) & 0.0249 & 0.0952 & 20.76 & 0.7591 & \textbf{0.0149} & \textbf{0.0682} & \textbf{24.26} & \textbf{0.8928} \\
    Traj.\ SSD (Rx) & 0.0766 & 0.1504 & 16.80 & 0.6703 & \textbf{0.0687} & \textbf{0.1398} & \textbf{17.55} & \textbf{0.7156} \\
    Traj.\ SSD (Tx) & 0.0530 & 0.1212 & 18.78 & 0.7047 & \textbf{0.0472} & \textbf{0.1115} & \textbf{19.65} & \textbf{0.7611} \\
    \midrule
    & \multicolumn{2}{c}{AoA Acc.$\uparrow$} & \multicolumn{2}{c}{AoD Acc.$\uparrow$} & \multicolumn{2}{c}{AoA Acc.$\uparrow$} & \multicolumn{2}{c}{AoD Acc.$\uparrow$} \\
    \cmidrule(lr){2-3}\cmidrule(lr){4-5}\cmidrule(lr){6-7}\cmidrule(l){8-9}
    Link Reasoning & \multicolumn{2}{c}{76.72\%} & \multicolumn{2}{c}{79.47\%} & \multicolumn{2}{c}{\textbf{89.95\%}} & \multicolumn{2}{c}{\textbf{92.84\%}} \\
    \bottomrule
  \end{tabular}
\end{table}

\paragraph{Pretrained World-Model Initialization.}
Tab.~\ref{tab:ablate-init} compares pretrained world model initialization against training from scratch under identical optimization budgets. Since RSSM synthesis is largely a 2D planar mapping governed by top-down geometry, dense spatial supervision from scratch could achieve good performance. In contrast, synthesizing directional SSDs and reasoning about propagation paths require inferring how EM waves interact with 3D scene geometry. Therefore, world-model pretraining provides a large performance gain for the SSD task and reasoning task, which shows that world-model pretraining effectively transfers spatial understanding priors to synthesize RF fields.

\section{Conclusion}
\label{sec:conclusion}
We present RadioWorld, a unified world-model framework that synthesizes multi-height RSSMs and bidirectional SSDs in a single deterministic forward pass without requiring any target-scene RF measurements. Evaluated on UrbanRF-Bench across seven global cities, RadioWorld substantially outperforms specialized baselines in unseen cities while demonstrating strong utility in downstream beam selection and base station deployment.

\paragraph{Limitations and Future Work.}
Despite strong zero-shot generalization across diverse urban layouts, RadioWorld currently relies on coarse 2.5D BEV occupancy and height maps under the static-scene assumption, focusing on power-domain RSSMs and SSDs. Leveraging the architectural flexibility and native multimodal reasoning capabilities of our dual-stream world model, these limitations point to two promising directions for future work. First, integrating 3D scene representations, such as point clouds and material-aware meshes, can capture fine-grained multipath information. Second, extending the framework from static urban layouts to dynamic environments with mobile transceivers will enable the synthesis of temporally evolving, dynamic RSSMs and SSDs. Looking ahead, we envision RadioWorld serving as a scalable foundation model for wireless physical AI.

\clearpage
\bibliographystyle{assets/plainnat}
\bibliography{paper}

@String(CVPR= {IEEE Conf. Comput. Vis. Pattern Recog.})

@String(ICCV= {Int. Conf. Comput. Vis.})

@String(ECCV= {Eur. Conf. Comput. Vis.})

@String(ICASSP=	{ICASSP})

@String(ICLR = {Int. Conf. Learn. Represent.})

@String(CVPR  = {CVPR})

@String(ICCV  = {ICCV})

@String(ECCV  = {ECCV})

@String(ICLR  = {ICLR})

@article{levie2021radiounet,
  title   = {{RadioUNet}: Fast Radio Map Estimation With Convolutional Neural Networks},
  author={Levie, Ron and Yapar, {\c{C}}a{\u{g}}kan and Kutyniok, Gitta and Caire, Giuseppe},
  journal = {IEEE Transactions on Wireless Communications},
  volume  = {20},
  number  = {6},
  pages   = {4001--4015},
  year    = {2021},
  publisher={IEEE}
}

@article{wang2025radiodiff,
  title   = {{RadioDiff}: An Effective Generative Diffusion Model for Sampling-Free Dynamic Radio Map Construction},
  author  = {Wang, Xiucheng and Tao, Keda and Cheng, Nan and Yin, Zhisheng and Li, Zan and Zhang, Yuan and Shen, Xuemin},
  journal = {IEEE Transactions on Cognitive Communications and Networking},
  volume  = {11},
  number  = {2},
  pages   = {738--750},
  year    = {2025},
  publisher={IEEE}
}

@inproceedings{zhao2023nerf2,
  title     = {{NeRF}$^2$: Neural Radio-Frequency Radiance Fields},
  author    = {Zhao, Xiaopeng and An, Zhenlin and Pan, Qingrui and Yang, Lei},
  booktitle = {Proceedings of the 29th Annual International Conference on Mobile Computing and Networking},
  series    = {MobiCom '23},
  pages     = {1--15},
  year      = {2023},
  publisher = {Association for Computing Machinery}
}

@inproceedings{wen2025wrfgs,
  title     = {{WRF-GS}: Wireless Radiation Field Reconstruction with {3D} {Gaussian} Splatting},
  author    = {Wen, Chaozheng and Tong, Jingwen and Hu, Yingdong and Lin, Zehong and Zhang, Jun},
  booktitle = {IEEE INFOCOM 2025 -- IEEE Conference on Computer Communications},
  pages     = {1--10},
  year      = {2025},
  organization={IEEE}
}

@inproceedings{yang2026graf,
  title     = {Generalizable Radio-Frequency Radiance Fields for Spatial Spectrum Synthesis},
  author    = {Yang, Kang and Chen, Yuning and Du, Wan},
  booktitle = {Proceedings of the IEEE/CVF Conference on Computer Vision and Pattern Recognition (CVPR)},
  pages={12533--12543},
  year      = {2026}
}

@inproceedings{hehn2025wigatr,
  title     = {Differentiable and Learnable Wireless Simulation with Geometric Transformers},
  author    ={Hehn, Thomas and Peschl, Markus and Orekondy, Tribhuvanesh and Behboodi, Arash and Brehmer, Johann},
  booktitle = {International Conference on Learning Representations (ICLR)},
  pages={45353--45373},
  year      = {2025},
}

@article{wen2026point2radio,
  title   = {{Point2Radio}: A Foundation Model for Cross-Scene Radio Fields from Material-Aware Point Clouds},
  author  = {Wen, Chaozheng and Bian, Chenghong and Chen, Hongze and Zhang, Jun},
  journal = {arXiv preprint arXiv:2607.28994},
  year    = {2026}
}

@article{nvidia2026cosmos3,
  title   = {Cosmos 3: Omnimodal World Models for Physical {AI}},
  author  ={Agarwal, Niket and Ali, Arslan and Allen, Jon and Antolini, Martin and Aubame, Adeline and Azzolini, Alisson and Bai, Junjie and Bala, Maciej and Balaji, Yogesh and Bapst, Josh and others},
  journal = {arXiv preprint arXiv:2606.02800},
  year    = {2026}
}

@article{ait_aoudia2025sionnart,
  title   = {{Sionna RT}: Technical Report},
  author={Aoudia, Fay{\c{c}}al A{\"\i}t and Hoydis, Jakob and Nimier-David, Merlin and Nicolet, Baptiste and Cammerer, Sebastian and Keller, Alexander},
  journal = {arXiv preprint arXiv:2504.21719},
  year    = {2025}
}

@article{tschannen2025siglip2,
  title         = {{SigLIP 2}: Multilingual Vision-Language Encoders with Improved Semantic Understanding, Localization, and Dense Features},
  author={Tschannen, Michael and Gritsenko, Alexey and Wang, Xiao and Naeem, Muhammad Ferjad and Alabdulmohsin, Ibrahim and Parthasarathy, Nikhil and Evans, Talfan and Beyer, Lucas and Xia, Ye and Mustafa, Basil and others},
  journal       = {arXiv preprint arXiv:2502.14786},
  year          = {2025}
}

@article{wang2025wan,
  title         = {Wan: Open and Advanced Large-Scale Video Generative Models},
  author={{Wan Team} and Wang, Ang and Ai, Baole and Wen, Bin and Mao, Chaojie and Xie, Chen-Wei and Chen, Di and Yu, Feiwu and Zhao, Haiming and Yang, Jianxiao and others},
  journal       = {arXiv preprint arXiv:2503.20314},
  year          = {2025}
}

@inproceedings{mur2026v,
  title={{V-JEPA} 2.1: Unlocking dense features in video self-supervised learning},
  author={Mur-Labadia, Lorenzo and Muckley, Matthew and Bar, Amir and Assran, Mido and Sinha, Koustuv and Rabbat, Mike and LeCun, Yann and Ballas, Nicolas},
  booktitle={European Conference on Computer Vision (ECCV)},
  pages={671--689},
  year={2026},
  organization={Springer}
}

@article{oquab2023dinov2,
  title={{DINOv2}: Learning robust visual features without supervision},
  author={Oquab, Maxime and Darcet, Timoth{\'e}e and Moutakanni, Th{\'e}o and Vo, Huy and Szafraniec, Marc and Khalidov, Vasil and Fernandez, Pierre and Haziza, Daniel and Massa, Francisco and El-Nouby, Alaaeldin and others},
  journal={arXiv preprint arXiv:2304.07193},
  year={2023}
}

@article{jia2025radiomamba,
  title   = {{RadioMamba}: Breaking the Accuracy-Efficiency Trade-Off in Radio Map Construction Via a Hybrid {Mamba-UNet}},
  author={Jia, Honggang and Cheng, Nan and Wang, Xiucheng and Zhou, Conghao and Sun, Ruijin and Shen, Xuemin},
  journal = {IEEE Transactions on Network Science and Engineering},
  volume={13},
  pages={2454--2468},
  year    = {2026}
}

@article{wen2025wrfgsplus,
  title   = {Neural Representation for Wireless Radiation Field Reconstruction: A {3D} {Gaussian} Splatting Approach},
  author  = {Wen, Chaozheng and Tong, Jingwen and Hu, Yingdong and Lin, Zehong and Zhang, Jun},
  journal = {IEEE Transactions on Wireless Communications},
  volume={25},
  pages={7490--7504},
  year    = {2025},
  publisher={IEEE}
}

@book{goldsmith2005wireless,
  title={Wireless Communications},
  author={Goldsmith, Andrea},
  publisher={Cambridge University Press},
  year={2005}
}

@article{yun2015ray,
  title   = {Ray Tracing for Radio Propagation Modeling: Principles and Applications},
  author={Yun, Zhengqing and Iskander, Magdy F},
  journal = {IEEE Access},
  volume  = {3},
  pages   = {1089--1100},
  year    = {2015}
}

@article{liang2024mot,
  title   = {Mixture-of-Transformers: A Sparse and Scalable Architecture for Multi-Modal Foundation Models},
  author={Liang, Weixin and Yu, Lili and Luo, Liang and Iyer, Srinivasan and Dong, Ning and Zhou, Chunting and Ghosh, Gargi and Lewis, Mike and Yih, Wen-tau and Zettlemoyer, Luke and others},
  journal = {arXiv preprint arXiv:2411.04996},
  year    = {2024}
}

@inproceedings{lipman2023flow,
  title     = {Flow Matching for Generative Modeling},
  author={Lipman, Yaron and Chen, Ricky TQ and Ben-Hamu, Heli and Nickel, Maximilian and Le, Matt},
  booktitle = {International Conference on Learning Representations (ICLR)},
  year      = {2023}
}

@inproceedings{hu2022lora,
  title     = {{LoRA}: Low-Rank Adaptation of Large Language Models},
  author    = {Hu, Edward J. and Shen, Yelong and Wallis, Phillip and Allen-Zhu, Zeyuan and Li, Yuanzhi and Wang, Shean and Wang, Lu and Chen, Weizhu},
  booktitle = {International Conference on Learning Representations (ICLR)},
  year      = {2022}
}

@inproceedings{lu2023newrf,
  title     = {{NeWRF}: A Deep Learning Framework for Wireless Radiation Field Reconstruction and Channel Prediction},
  author={Lu, Haofan and Vattheuer, Christopher and Mirzasoleiman, Baharan and Abari, Omid},
  booktitle = {International Conference on Machine Learning (ICML)},
  year      = {2024}
}

@article{romero2022radiomap,
  author  = {Romero, Daniel and Kim, Seung-Jun},
  title   = {Radio Map Estimation: A Data-Driven Approach to Spectrum Cartography},
  journal = {IEEE Signal Processing Magazine},
  volume  = {39},
  number  = {6},
  pages   = {53--72},
  year    = {2022}
}

@article{zhang2023rmegan,
  title   = {{RME-GAN}: A Learning Framework for Radio Map Estimation Based on Conditional Generative Adversarial Network},
  author={Zhang, Songyang and Wijesinghe, Achintha and Ding, Zhi},
  journal = {IEEE Internet of Things Journal},
  volume  = {10},
  number  = {20},
  pages   = {18016--18027},
  year    = {2023}
}

@inproceedings{ke2024marigold,
  title     = {Repurposing Diffusion-Based Image Generators for Monocular Depth Estimation},
    author={Ke, Bingxin and Obukhov, Anton and Huang, Shengyu and Metzger, Nando and Daudt, Rodrigo Caye and Schindler, Konrad},
  booktitle = {Proceedings of the IEEE/CVF Conference on Computer Vision and Pattern Recognition (CVPR)},
  pages={9492--9502},
  year={2024},
}

@inproceedings{garcia2025finetuning,
  title     = {Fine-Tuning Image-Conditional Diffusion Models is Easier than You Think},
    author={Garcia, Gonzalo Martin and Zeid, Karim Abou and Schmidt, Christian and De Geus, Daan and Hermans, Alexander and Leibe, Bastian},
  booktitle = {IEEE/CVF Winter Conference on Applications of Computer Vision (WACV)},
  pages={753--762},
  year      = {2025}
}

@inproceedings{he2025lotus,
  title     = {Lotus: Diffusion-Based Visual Foundation Model for High-Quality Dense Prediction},
  author={He, Jing and Li, Haodong and Yin, Wei and Liang, Yixun and Li, Leheng and Zhou, Kaiqiang and Zhang, Hongbo and Liu, Bingbing and Chen, YingCong},
  booktitle = {International Conference on Learning Representations (ICLR)},
  pages={89454--89467},
  year      = {2025}
}

@inproceedings{bruce2024genie,
  title     = {Genie: Generative Interactive Environments},
  author={Bruce, Jake and Dennis, Michael D and Edwards, Ashley and Parker-Holder, Jack and Shi, Yuge and Hughes, Edward and Lai, Matthew and Mavalankar, Aditi and Steigerwald, Richie and Apps, Chris and others},
  booktitle = {International Conference on Machine Learning (ICML)},
  year      = {2024}
}

@misc{brooks2024sora,
  author={Brooks, Tim and Peebles, Bill and Holmes, Connor and DePue, Will and Guo, Yufei and Jing, Leo and Schnurr, David and Taylor, Joe and Luhman, Troy and Luhman, Eric and others},
  title        = {Video Generation Models as World Simulators},
  howpublished = {OpenAI technical report},
  year         = {2024},
  url          = {https://openai.com/index/video-generation-models-as-world-simulators/}
}

@article{nvidia2025cosmos,
  title   = {Cosmos World Foundation Model Platform for Physical {AI}},
  author={Agarwal, Niket and Ali, Arslan and Bala, Maciej and Balaji, Yogesh and Barker, Erik and Cai, Tiffany and Chattopadhyay, Prithvijit and Chen, Yongxin and Cui, Yin and Ding, Yifan and others},
  journal = {arXiv preprint arXiv:2501.03575},
  year    = {2025}
}

@article{assran2025vjepa2,
  title   = {{V-JEPA} 2: Self-Supervised Video Models Enable Understanding, Prediction and Planning},
  author={Assran, Mido and Bardes, Adrien and Fan, David and Garrido, Quentin and Howes, Russell and Muckley, Matthew and Rizvi, Ammar and Roberts, Claire and Sinha, Koustuv and Zholus, Artem and others},
  journal = {arXiv preprint arXiv:2506.09985},
  year    = {2025}
}

@inproceedings{yang2025gsrf,
  title     = {{GSRF}: Complex-Valued {3D} {Gaussian} Splatting for Efficient Radio-Frequency Data Synthesis},
  author={Yang, Kang and Dong, Gaofeng and Ji, Sijie and Du, Wan and Srivastava, Mani},
  booktitle = {Advances in Neural Information Processing Systems (NeurIPS)},
  volume={38},
  pages={63076--63105},
  year={2025}
}

@article{giordani2019beam,
  title   = {A Tutorial on Beam Management for {3GPP} {NR} at {mmWave} Frequencies},
  author={Giordani, Marco and Polese, Michele and Roy, Arnab and Castor, Douglas and Zorzi, Michele},
  journal = {IEEE Communications Surveys \& Tutorials},
  volume  = {21},
  number  = {1},
  pages   = {173--196},
  year    = {2019}
}

@article{hoydis2022sionna,
  title   = {Sionna: An Open-Source Library for Next-Generation Physical Layer Research},
  author={Hoydis, Jakob and Cammerer, Sebastian and Aoudia, Fay{\c{c}}al A{\"\i}t and Vem, Avinash and Binder, Nikolaus and Marcus, Guillermo and Keller, Alexander},
  journal = {arXiv preprint arXiv:2203.11854},
  year    = {2022}
}

@book{vantrees2002optimum,
  author    = {Van Trees, Harry L.},
  title     = {Optimum Array Processing: Part {IV} of Detection, Estimation, and Modulation Theory},
  publisher={John Wiley \& Sons},
  year      = {2002}
}

@article{heath2016overview,
  title   = {An Overview of Signal Processing Techniques for Millimeter Wave {MIMO} Systems},
  author={Heath, Robert W and Gonzalez-Prelcic, Nuria and Rangan, Sundeep and Roh, Wonil and Sayeed, Akbar M},
  journal = {IEEE Journal of Selected Topics in Signal Processing},
  volume  = {10},
  number  = {3},
  pages   = {436--453},
  year    = {2016}
}

@inproceedings{yapar2023challenge,
  title     = {The First Pathloss Radio Map Prediction Challenge},
  author={Yapar, {\c{C}}a{\u{g}}kan and Jaensch, Fabian and Levie, Ron and Kutyniok, Gitta and Caire, Giuseppe},
  booktitle = {IEEE International Conference on Acoustics, Speech and Signal Processing (ICASSP)},
  year      = {2023}
}

@inproceedings{lee2023pmnet,
  title     = {{PMNet}: Robust Pathloss Map Prediction via Supervised Learning},
  author={Lee, Ju-Hyung and Serbetci, Omer Gokalp and Selvam, Dheeraj Panneer and Molisch, Andreas F},
  booktitle = {IEEE Global Communications Conference (GLOBECOM)},
  year      = {2023}
}

@article{mildenhall2020nerf,
  title={{NeRF}: Representing scenes as neural radiance fields for view synthesis},
  author={Mildenhall, Ben and Srinivasan, Pratul P and Tancik, Matthew and Barron, Jonathan T and Ramamoorthi, Ravi and Ng, Ren},
  journal={Communications of the ACM},
  volume={65},
  number={1},
  pages={99--106},
  year={2022},
  publisher={ACM, New York, NY, USA}
}

@article{kerbl2023gaussian,
  title   = {{3D} {Gaussian} Splatting for Real-Time Radiance Field Rendering},
  author={Kerbl, Bernhard and Kopanas, Georgios and Leimk{\"u}hler, Thomas and Drettakis, George and others},
  journal = {ACM Transactions on Graphics},
  volume  = {42},
  number  = {4},
  year    = {2023}
}

@inproceedings{orekondy2023winert,
  title     = {{WiNeRT}: Towards Neural Ray Tracing for Wireless Channel Modelling and Differentiable Simulations},
  author={Orekondy, Tribhuvanesh and Kumar, Pratik and Kadambi, Shreya and Ye, Hao and Soriaga, Joseph and Behboodi, Arash},
  booktitle = {International Conference on Learning Representations (ICLR)},
  year      = {2023}
}

@article{simeoni2025dinov3,
  title   = {{DINOv3}},
  author={Sim{\'e}oni, Oriane and Vo, Huy V and Seitzer, Maximilian and Baldassarre, Federico and Oquab, Maxime and Jose, Cijo and Khalidov, Vasil and Szafraniec, Marc and Yi, Seungeun and Ramamonjisoa, Micha{\"e}l and others},
  journal = {arXiv preprint arXiv:2508.10104},
  year    = {2025}
}

@inproceedings{fu2024geowizard,
  title     = {{GeoWizard}: Unleashing the Diffusion Priors for {3D} Geometry Estimation from a Single Image},
  author={Fu, Xiao and Yin, Wei and Hu, Mu and Wang, Kaixuan and Ma, Yuexin and Tan, Ping and Shen, Shaojie and Lin, Dahua and Long, Xiaoxiao},
  booktitle = {European Conference on Computer Vision (ECCV)},
  pages={241--258},
  year      = {2024}
}

@article{hafner2023dreamerv3,
  title   = {Mastering Diverse Domains through World Models},
  author={Hafner, Danijar and Pasukonis, Jurgis and Ba, Jimmy and Lillicrap, Timothy},
  journal = {arXiv preprint arXiv:2301.04104},
  year    = {2023}
}

@inproceedings{yang2024unisim,
  title     = {Learning Interactive Real-World Simulators},  
  author    = {Yang, Sherry and Du, Yilun and Ghasemipour, Seyed Kamyar Seyed and Tompson, Jonathan and Kaelbling, Leslie and Schuurmans, Dale and Abbeel, Pieter},
  booktitle = {International Conference on Learning Representations (ICLR)},
  year      = {2024}
}

@inproceedings{bar2025nwm,
  author    = {Bar, Amir and Zhou, Gaoyue and Tran, Danny and Darrell, Trevor and LeCun, Yann},
  title     = {Navigation World Models},
  booktitle = {Proceedings of the IEEE/CVF Conference on Computer Vision and Pattern Recognition (CVPR)},
  pages={15791--15801},
  year      = {2025}
}

@inproceedings{peebles2023dit,
  author    = {Peebles, William and Xie, Saining},
  title     = {Scalable Diffusion Models with Transformers},
  booktitle = {International Conference on Computer Vision (ICCV)},
  pages={4172--4182},
  year      = {2023}
}

@inproceedings{liu2023rectified,
  author    = {Liu, Xingchao and Gong, Chengyue and Liu, Qiang},
  title     = {Flow Straight and Fast: Learning to Generate and Transfer Data with Rectified Flow},
  booktitle = {International Conference on Learning Representations (ICLR)},
  year      = {2023}
}

@inproceedings{tancik2020fourier,
  title     = {Fourier Features Let Networks Learn High Frequency Functions in Low Dimensional Domains},
  author={Tancik, Matthew and Srinivasan, Pratul and Mildenhall, Ben and Fridovich-Keil, Sara and Raghavan, Nithin and Singhal, Utkarsh and Ramamoorthi, Ravi and Barron, Jonathan and Ng, Ren},
  booktitle = {Advances in Neural Information Processing Systems (NeurIPS)},
  volume={33},
  pages={7537--7547},
  year      = {2020}
}

@inproceedings{loshchilov2019adamw,
  author    = {Loshchilov, Ilya and Hutter, Frank},
  title     = {Decoupled Weight Decay Regularization},
  booktitle = {International Conference on Learning Representations (ICLR)},
  year      = {2019}
}

@article{wang2004ssim,
  author  = {Wang, Zhou and Bovik, Alan C. and Sheikh, Hamid R. and Simoncelli, Eero P.},
  title   = {Image Quality Assessment: From Error Visibility to Structural Similarity},
  journal = {IEEE Transactions on Image Processing},
  volume  = {13},
  number  = {4},
  pages   = {600--612},
  year    = {2004}
}

@inproceedings{
zhu2026video,
title={Video Generation Empowered Long-Term Radio Map Prediction in {UAV}-Assisted Communication},
author={Lin Zhu and Weifeng Zhu and Shuowen Zhang and Liang Liu},
booktitle={Second Workshop of AI4NextG @ ICML 2026: AI and ML for Next-Generation Wireless},
year={2026}
}

@article{zhu2026df,
  title={{DF-3DRME}: A Data-Friendly Learning Framework for {3D} Radio Map Estimation based on Super-Resolution Technique},
  author={Zhu, Lin and Zhu, Weifeng and Zhang, Shuowen and Caire, Giuseppe and Liu, Liang},
  journal={arXiv preprint arXiv:2604.00676},
  year={2026}
}

\end{document}